\documentclass[10pt,twocolumn,letterpaper]{article}

\usepackage[pagenumbers]{cvpr} 

\usepackage{graphicx}
\usepackage{amsmath}
\usepackage{amssymb}
\usepackage{booktabs}
\usepackage{multirow}
\usepackage{multicol}
\usepackage{pifont}
\newcommand{\cmark}{\textcolor{green}{\ding{51}}}%
\newcommand{\xmark}{\textcolor{red}{\ding{55}}}%

\usepackage[pagebackref,breaklinks,colorlinks]{hyperref}

\usepackage[capitalize]{cleveref}
\crefname{section}{Sec.}{Secs.}
\Crefname{section}{Section}{Sections}
\Crefname{table}{Table}{Tables}
\crefname{table}{Tab.}{Tabs.}

\def\cvprPaperID{7833} 
\def\confName{CVPR}
\def\confYear{2023}

\begin{document}

\title{
Pushing the Limits of Asynchronous Graph-based Object Detection \\with Event Cameras
}

\makeatletter
\let\@oldmaketitle\@maketitle
\renewcommand{\@maketitle}{\@oldmaketitle
\centering
\setlength{\tabcolsep}{2pt} 
\begin{tabular}{cccc}
\includegraphics[height=0.2\linewidth]{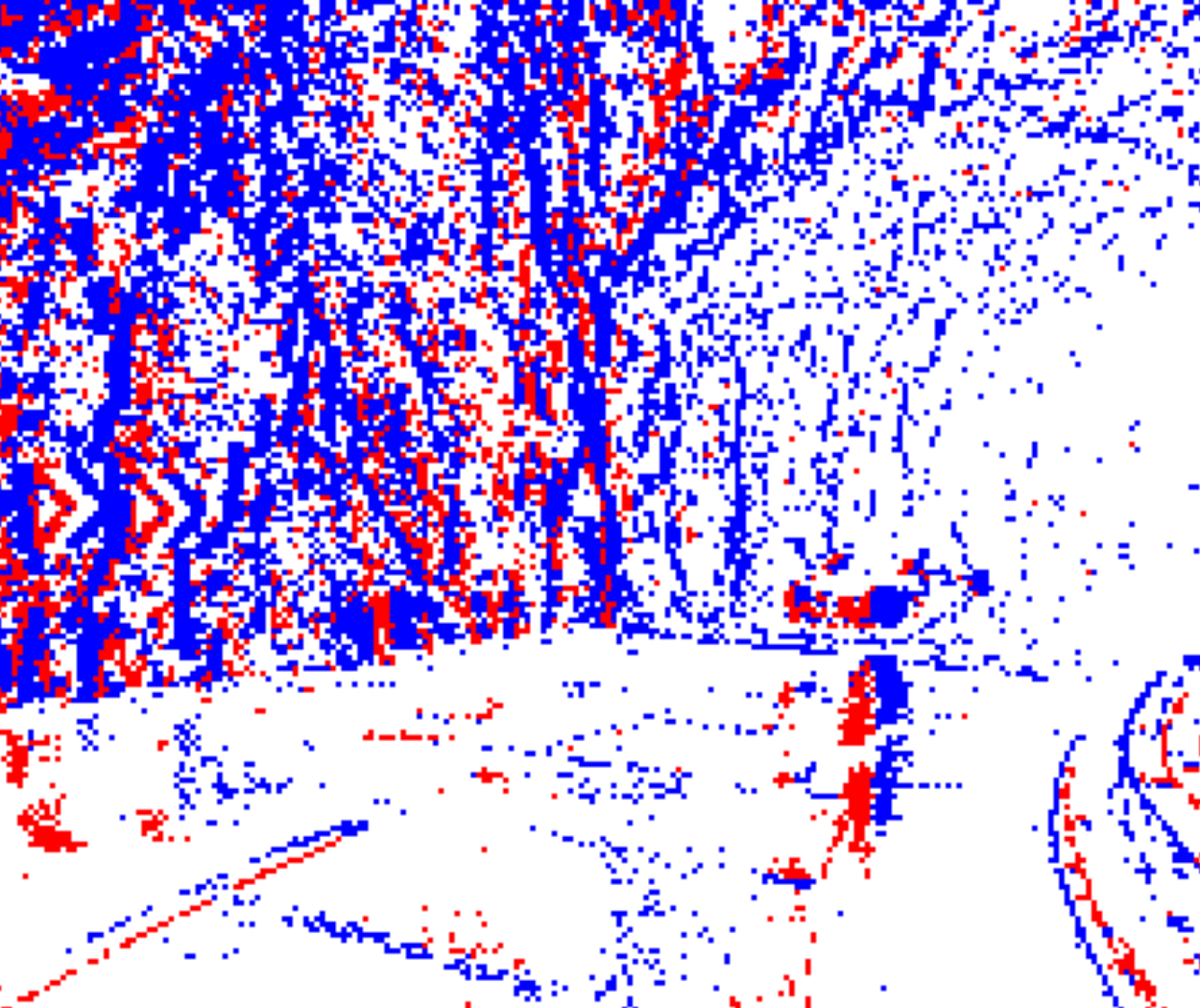}&
\includegraphics[height=0.2\linewidth]{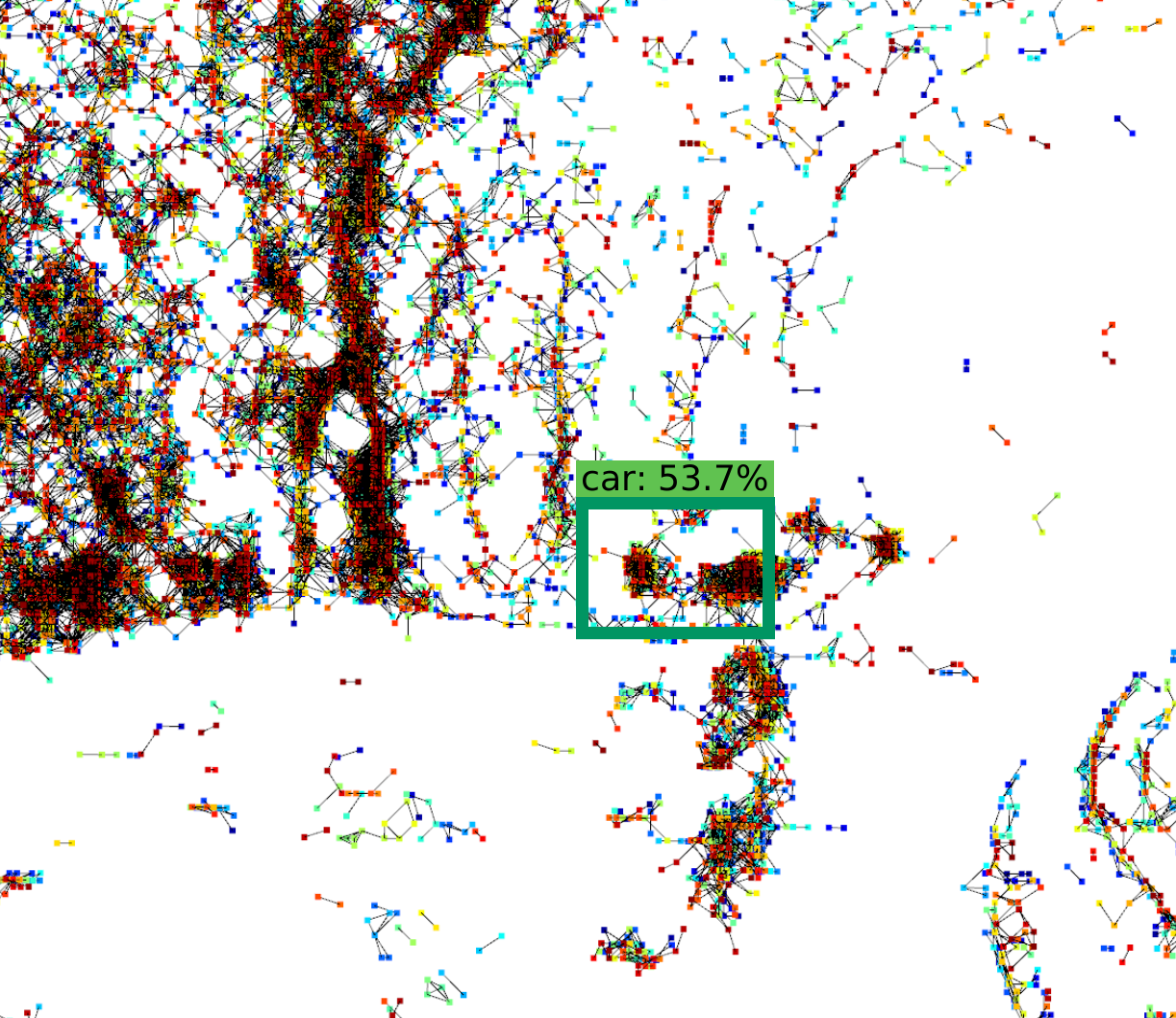}&
\includegraphics[height=0.2\linewidth]{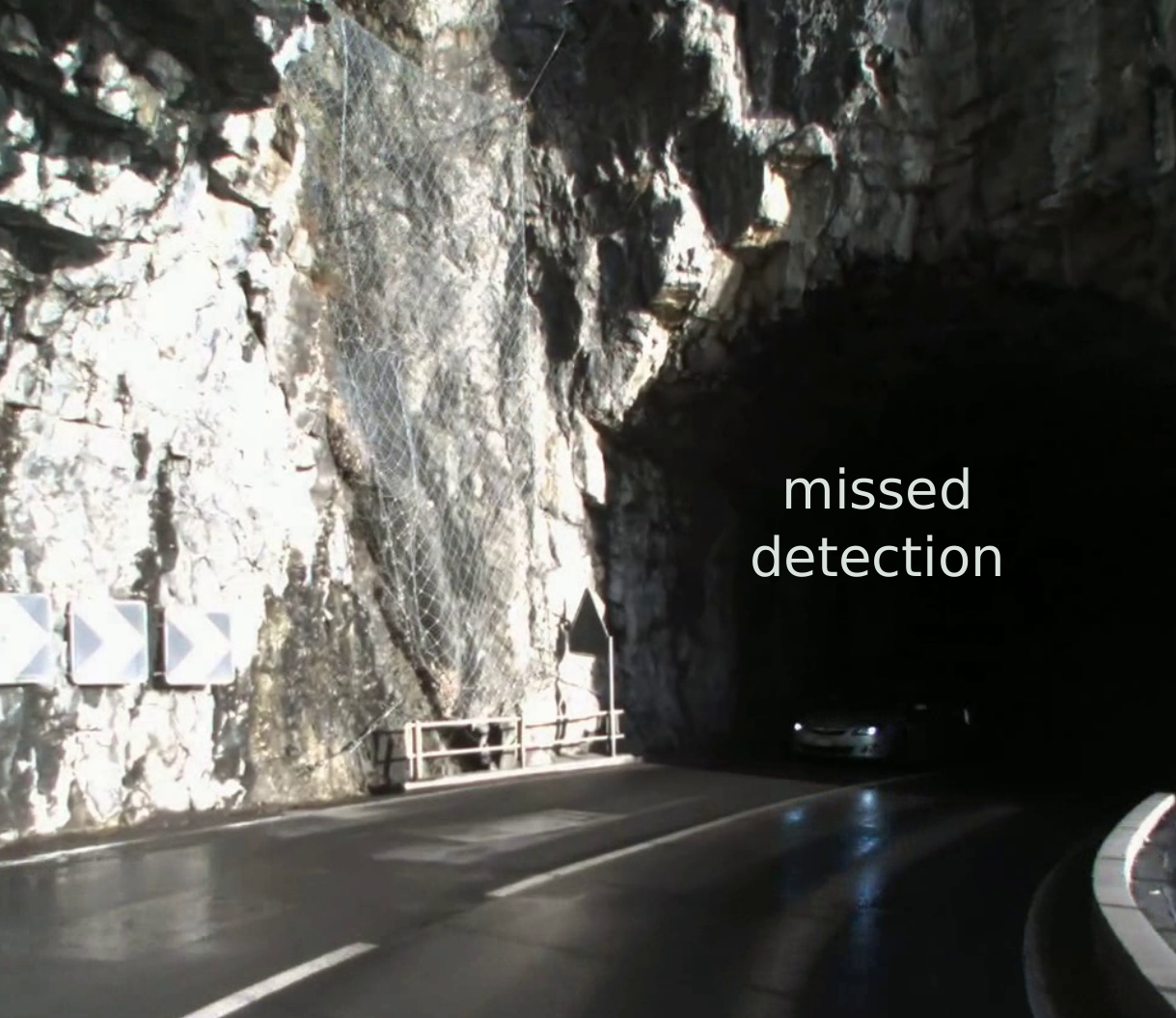}&
\includegraphics[height=0.2\linewidth]{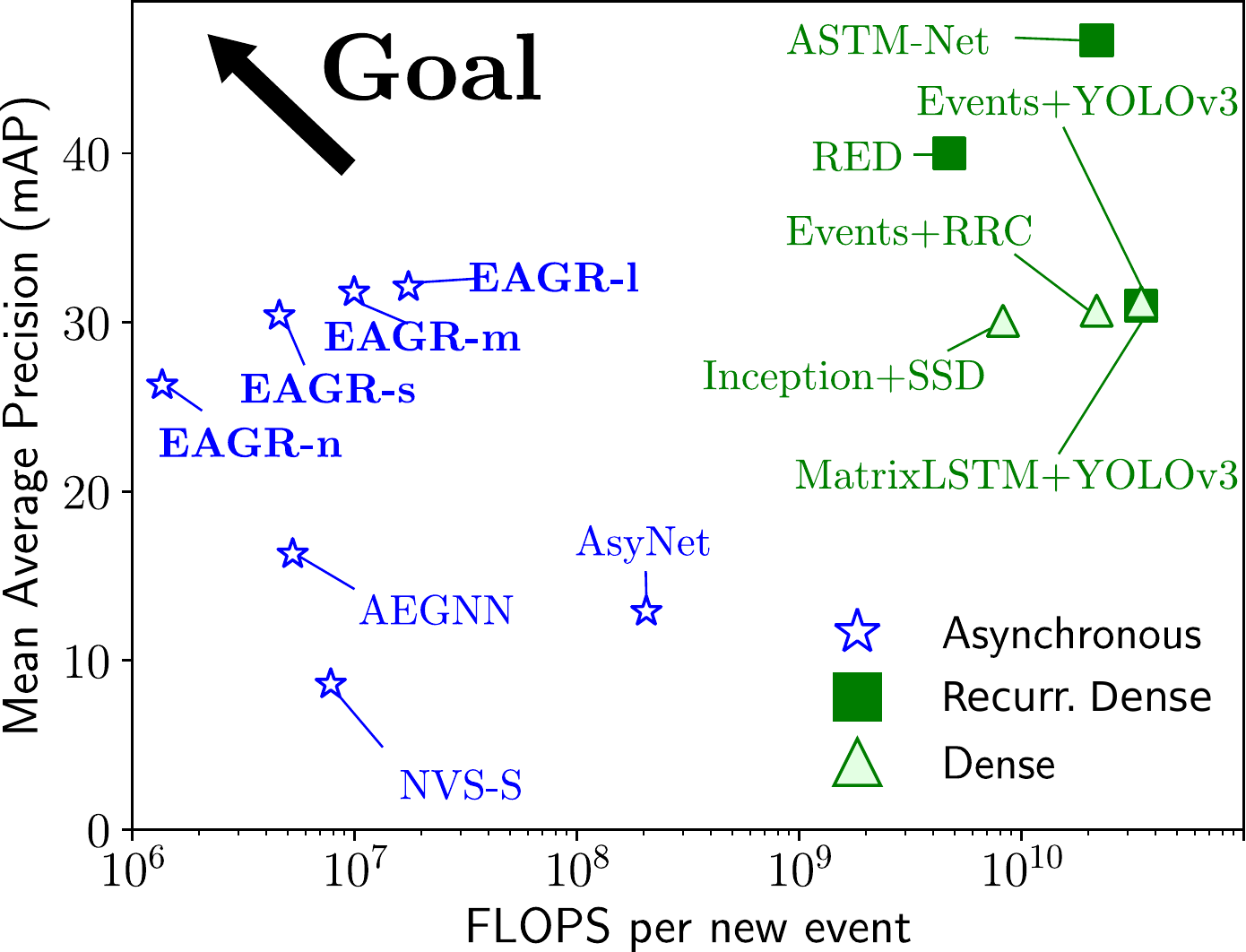}\\
(a) Raw events& (b) Graph-based detection& (c) YOLOX~\cite{Ge21arxiv} on image & (d) Efficiency vs. Performance
\end{tabular}

\vspace{-1ex}
\captionof{figure}{We introduce a new class of event-based object detectors, which we term \textbf{E}fficient \textbf{A}synchronous \textbf{Gr}aph Neural Networks (\textbf{EAGR}). They process events (a) as spatiotemporally evolving event graphs (b) and can be deployed in an efficient, asynchronous mode where they only perform local updates for each new event, thus significantly reducing computational complexity. EAGRs have a two times deeper architecture compared with other asynchronous GNNs~\cite{Schaefer22cvpr,Li21iccv} and are more efficient in event-by-event processing (d, blue box).
This opens the door to efficient, and accurate object detection in edge-case scenarios (c) (from~\cite{Gehrig21ral}), where methods like YOLOX-s~\cite{Ge21arxiv} based on standard images fail.
\vspace{0.3cm}}
\label{fig:eyecatcher}
}
\makeatother
\setlength{\tabcolsep}{10pt} 

\author{Daniel Gehrig and Davide Scaramuzza \\\\
Robotics and Perception Group, University of Zurich
}
\maketitle

\begin{abstract}
\vspace{-2ex}
   State-of-the-art machine-learning methods for event cameras treat events as dense representations and process them with conventional deep neural networks. Thus, they fail to maintain the sparsity and asynchronous nature of event data, thereby imposing significant computation and latency constraints on downstream systems. A recent line of work tackles this issue by modeling events as spatiotemporally evolving graphs that can be efficiently and asynchronously processed using graph neural networks. These works showed impressive computation reductions, yet their accuracy is still limited by the small scale and shallow depth of their network, both of which are required to reduce computation. In this work, we break this glass ceiling by introducing several architecture choices which allow us to scale the depth and complexity of such models while maintaining low computation. On object detection tasks, our smallest model shows up to 3.7 times lower computation, while outperforming state-of-the-art asynchronous methods by 7.4 mAP. Even when scaling to larger model sizes, we are 13\% more efficient than state-of-the-art while outperforming it by 11.5 mAP. As a result, our method runs 3.7 times faster than a dense graph neural network, taking only 8.4 ms per forward pass. 
   This opens the door to efficient, and accurate object detection in edge-case scenarios.
\end{abstract}
\vspace{-4ex}
\section{Introduction}
\label{sec:intro}
Humans can detect fast-moving objects in the blink of an eye thanks to their sophisticated visual cortex originally designed to hunt and spot prey. 
Today, computer vision researchers try to emulate these systems with data-driven object detection algorithms, which have found widespread application in robotic and automotive settings. 
However, state-of-the-art approaches operate on data from frame-based sensors like RGB cameras or LiDARs and, for this reason, suffer from a bandwidth-latency tradeoff:
at high speeds, they require a high framerate to reduce perceptual latency, but this introduces a significant bandwidth overhead for downstream systems; 
reducing the framerate reduces the bandwidth requirements but at the cost of missing important scene dynamics, like the ones in Fig.~\ref{fig:eyecatcher}, due to increased perceptual latency.   

In recent years, event cameras have emerged as alternative sensors that do not suffer from this tradeoff: they are bio-inspired vision sensors that only measure \emph{changes in intensity} that exceed a given threshold. 
These changes, called \emph{events}, are recorded asynchronously and with microsecond resolution. 
Due to their working principle, event cameras output sparse data with only a fraction of the bandwidth and power used by conventional RGB cameras. Also, they can adapt to the scene dynamics, featuring sub-millisecond perceptual latency at all speeds~\cite{Lichtsteiner08ssc,Brandli14ssc}. 
For an overview of applications and methods for event-based vision see~\cite{Gallego20pami}.

Despite this promise, state-of-the-art, event-based object detectors still do not leverage the event sparsity and instead, convert them into dense frame-like representations~\cite{Tulyakov19iccv,Tulyakov21cvpr,Gehrig19iccv,Zhu19cvpr,Rebecq19cvpr,Iacono18iros,Jiang19icra,Li22trip,Perot20nips,Cannici20eccv}. These are then processed with deep Convolutional Neural Networks (CNNs) originally designed to work well for standard images. However, they require a significant amount of computation, most of which is redundant or operates on artificial zeros. 

A recent line of work has tried to bring back efficient computation to dense methods by modeling events as spatio-temporal spike trains~\cite{Cordone22ijcnn}, point clouds~\cite{Sekikawa19cvpr}, or graphs~\cite{Li21iccv,Schaefer22cvpr,Bi19iccv,Deng22cvpr}, and processing them with corresponding specialized neural network architectures. Among these, graph neural networks (GNNs) have shown the highest efficiency and performance promise, leveraging insights from the fast-growing field of deep learning on graphs. 
The works in~\cite{Li21iccv,Schaefer22cvpr} have shown that once trained, such GNNs can be deployed in an asynchronous \emph{event-by-event} processing mode with identical output. In this mode, only local updates are performed for each new event, which are propagated to deeper layers. By limiting the computation to local subgraphs, these methods reduce the processing compared to dense methods by efficiently reusing past computations.

Despite these gains in efficiency compared to dense processing methods, graph neural networks trained on events are still behind in terms of expressiveness and accuracy. 
This is because current asynchronous GNNs are artificially kept shallow and lightweight to limit the per-event computation. Per-event computation still scales with the feature dimension and size of the subgraphs, which grow as the depth of the network increases. Thus, asynchronous methods hit a glass ceiling, since, as network depth increases their benefits over synchronous processing diminish.

In this work, we introduce several architecture design choices that allow us to scale the depth and capacity of GNNs significantly while maintaining highly efficient per-event processing.
To maintain a low computational complexity, we first investigate effective ways to \emph{prune node updates} caused by new events, so that they do not need to be propagated to lower layers. We find that, when combined with max pooling in early layers and node position rounding, we can skip up to 73\% of the computation in lower layers with a minimal performance impact. As a next step, we show the significant benefit of performing \emph{early temporal node aggregation}, which simultaneously boosts performance by 10.6 mAP and allows us to deploy our networks with novel and efficient Look-up-based Spline Convolutions (LUT-SCs). These LUT-SCs are trained as regular Spline Convolutions~\cite{Fey2018cvpr}, but are later deployed as efficient look-up tables, which require 4.5 times less computation. Finally, we leverage directed event graphs (DEGs) at the input to boost our model's performance by 1.8 mAP with a minimal computational cost. This is because, the k-hop subgraphs of DEGS maintain a \emph{constant} size, thereby limiting their growth in lower layers. In summary: 
\begin{itemize}
    \item We introduce architecture designs to scale the depth and capacity of asynchronous graph neural networks while maintaining highly efficient processing. We achieve this by leveraging node update pruning, early temporal aggregation through max pooling, novel efficient Look-Up-Spline Convolutions, and directed event graph processing in early layers. Cumulatively, these factors reduce the computation by a factor of 33.1, while boosting the performance by 10.6 mAP. 
    \item We introduce several event-based object detectors, which follow these design principles, and come in four sizes: nano, small, medium, and large. Our nano model uses 3.8 times less computation while outperforming the most accurate asynchronous methods by 7.4 mAP. Simultaneously, our medium model still has a 13\% lower computational complexity than the most efficient method, while outperforming it by 11.5 mAP.
    \item We show that our small model performs on par with dense feed-forward neural networks while outperforming state-of-the-art asynchronous methods in terms of efficiency. Finally, our large model outperforms all feed-forward dense methods, paving the way for both accurate and efficient event-based object detection. 
\end{itemize}
\begin{figure*}[t!]
    \centering
    \vspace{-1ex}
    \includegraphics[width=0.9\linewidth]{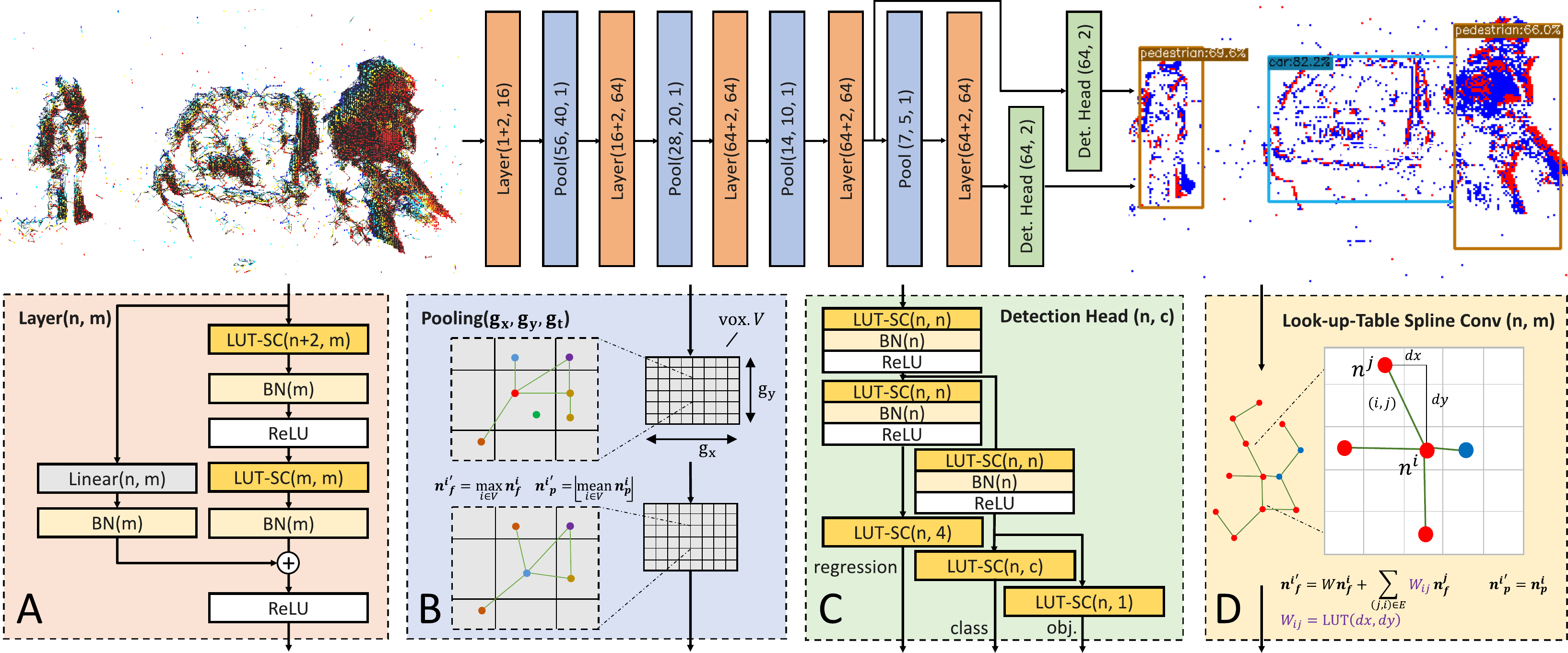}
    \caption{Overview of the network architecture of EAGR. It comprises a series of residual blocks followed by max pooling layers. For the residual blocks, the arguments $n$ and $m$ denote input an output channels dimension. The $+2$ indicates that we concatenate the 2D node position at that scale before processing. For pooling arguments $g_x$, $g_y$ and $g_t$ denote the number of grid cells in each dimension. Furthermore, we use a multiscale YOLOX-inspired detection head, outputting bounding boxes (\emph{regression}), class scores and object confidence. The basic building block is the Look-up-Table Spline Convolution (LUT-SC). It uses the discrete-valued relative distance between neighboring nodes to look up a weight matrix which is uses to compute the message sent to the center node.}
    \label{fig:method_overview}
    \vspace{-2ex}
\end{figure*}

\section{Related Work}
Since the introduction of powerful object detectors in classical image-based computer vision, such as R-CNN~\cite{Girshick14cvpr,Girshick15iccv,Ren15neurips}, SSD~\cite{Liu16eccv} and the YOLO series\cite{Redmon16cvpr, Ge21arxiv, Redmon18arxiv}, event-based object detection research has focused on leveraging the available models on dense, "image-like" event representations~\cite{Iacono18iros,Jiang19icra,Perot20nips,Cannici20eccv,Chen18cvprw,Li22trip}. This approach allows to use pretraining, and well-established architecture designs and loss functions, while maintaining the advantages of events, such as a high-dynamic range, and negligible motion blur. 

Most recent examples of such methods include RED~\cite{Perot20nips} and ASTMNet~\cite{Li22trip} which operate recurrently on events and have shown high performance on detection tasks in automotive settings. However, due to the nature of their method, these approaches necessarily need to convert events into dense frames. This invariably sacrifices the efficiency and high temporal resolution present in the events, which are critical in many application scenarios such as low-power, always-on, surveillance \cite{Indiveri11fns,Mitra09tbcs}, and low-latency, low-power object detection and avoidance~\cite{Falanga20science,Sanket20icra} . 

As a result, a parallel line of research has emerged which tries to reintroduce sparsity into the present models by either adopting spiking neural network architectures~\cite{Cordone22ijcnn} or geometric learning approaches~\cite{Messikommer20eccv,Schaefer22cvpr}. Of these, spiking neural networks are capable of processing raw events asynchronously and are thus closest in spirit to event-based data. However, they lack efficient learning rules and thus do not yet scale to complex tasks and datasets~\cite{Gehrig20icra,Lee16fns,Orchard15fns,Amir17cvpr,PerezCarrasco13pami,Sironi18cvpr}. Recently, geometric learning approaches have filled this gap. They treat events as spatio-temporal point-clouds~\cite{Sekikawa19cvpr}, submanifolds~\cite{Messikommer20eccv} or graphs~\cite{Li21iccv,Bi19iccv,Schaefer22cvpr,Mitrokhin18iros}, and process them with specialized neural networks. These methods retain the spatio-temporal sparsity in the events and can be implemented recursively, where single event insertions are highly efficient. 
Of these, processing events with graph-neural networks has proven to be most scalable, achieving high performance on complex tasks like object recognition \cite{Bi19iccv,Li21iccv,Deng22cvpr}, object detection \cite{Schaefer22cvpr} and motion segmentation~\cite{Mitrokhin20cvpr}. Simultaneously, they can be updated efficiently and asynchronously, for each new event, by only limiting computation to locally changed subgraphs, and propagating these changes to deeper layers in the network.

However, they are still far from achieving the same level of accuracy as dense methods. Due to efficiency requirements, current asynchronous graph neural networks are limited in terms of capacity and depth of the networks~\cite{Schaefer22cvpr,Li21iccv,Deng22cvpr}. This is because asynchronous methods become less efficient as the depth of the network increases, and their complexity still scales with the number of network parameters. In this work, we address this limitation by introducing a class of graph neural networks, which simultaneously has deep and high-capacity networks, but only has low computational complexity.

\section{Method}
In Sec. \ref{sec:event_graph} we start by reviewing the generation and data structure of events, followed by a description of how to create a graph from events. Next, in Sec.~\ref{sec:network_overview} we introduce our neural network, and finally describe the proposed asynchronous operation in Sec.~\ref{sec:asynchronous_operation}.  

\subsection{Event Generation and Graph Construction}
\label{sec:event_graph}
Event cameras have independent pixels which respond asynchronously to changes in logarithmic brightness $\mathbf{L}$. Whenever the magnitude of this change exceeds the contrast threshold $C$, that pixel triggers an event $e_i=(\mathbf{x}_i, t_i, p_i)$ characterized by the position $\mathbf{x}_i$, timestamp $t_i$ with microsecond resolution and polarity (sign) $p_i\in\{-1,1\}$ of the change. An event is triggered when 
\begin{equation}
    \label{eq:generative_model}
    p_i [\mathbf{L}(\mathbf{x}_i, t_i) - \mathbf{L}(\mathbf{x}_i, t_i-\Delta t_i)] > C.
\end{equation}
The event camera thus outputs a sparse stream of events $\mathcal{E}=\{e_i\}_{i=0}^{N-1}$. 
As in~\cite{Bi19iccv,Deng22cvpr,Schaefer22cvpr,Mitrokhin20cvpr,Li21iccv}, we interpret events as a 3D point cloud, connected via spatio-temporal edges. \\

Our event graph $\mathcal{G}=\{\mathcal{V}, E\}$ consists of nodes $\mathcal{V}$ and edges $E$. Each event $e_i$ corresponds to a node. These nodes $\mathbf{n}^i\in\mathcal{V}$ are characterized by their position $\textbf{n}^i_p=(\hat{\textbf{x}}_i, \beta t_i)\in \mathbb{R}^3$ and node features $\textbf{n}^i_f=p_i \in\mathbb{R}$. Here $\hat{\textbf{x}}_i$ is the event pixel coordinate, normalized by the height and width, and $t_i$ and $p_i$ are taken from the corresponding event. To map $t_i$ into the same range as $\textbf{x}_i$ we rescale it by a factor of $\beta=10^{-6}$. These nodes are connected via edges, $(i,j)\in E$, connecting nodes $\textbf{n}_i$ and $\textbf{n}_j$, each with edge attributes $e_{ij}\in \mathbb{R}^{d_e}$.
We connect nodes that are within a spatio-temporal distance from each other and temporally ordered
\begin{equation}
    \label{eq:edge_contraint}
    (i,j) \in E \quad\text{if}\quad   \Vert \mathbf{n}^i_p - \mathbf{n}^j_p \Vert_\infty < R \text{ and } t_i < t_j
\end{equation}
Here $\Vert .\Vert_\infty$ denotes the Manhattan distance, which returns the absolute value of the largest component.  Constructing the graph in this way gives us several advantages: First, we can leverage the queue-based graph construction method in \cite{Li21iccv} to implement a highly parallel graph construction algorithm on GPU. Our implementation constructs full event graphs with 50'000 nodes in 1.75 ms, and inserts single events in 0.3 ms on a Quadro RTX 4000 laptop GPU. 
Secondly, the temporal ordering constraint above, makes the event graph directed~\cite{Mitrokhin20cvpr, Li21iccv} which will enable high efficiency in early layers before pooling (see Sec.~\ref{sec:asynchronous_operation}). In this work, we select $R=0.01$ and limit the number of neighbors of each node to 16.

\subsection{Efficient Asynchronous Graph Neural Network}
\label{sec:network_overview}
An overview of our neural network architecture, which we term \textbf{E}fficient \textbf{A}synchronous \textbf{Gr}aph Neural Network, is shown in Fig.~\ref{fig:method_overview}. It processes the spatio-temporal graphs from Sec.~\ref{sec:event_graph} (Fig.~\ref{fig:method_overview}, top left) and outputs object detection at multiple scales (top right). It consists of five alternating residual layers (Fig.~\ref{fig:method_overview} A) and max pooling blocks (Fig.~\ref{fig:method_overview} B), followed by a YOLOX-inspired detection head at two scales (Fig.~\ref{fig:method_overview} C). 
Crucially, our network has a total of 13 convolution layers. By contrast, the methods in \cite{Li21iccv} and \cite{Schaefer22cvpr} feature only 5 and 7 layers respectively, making our network almost twice as deep as previous methods. 
Before each residual layer, we concatenate the $x$ and $y$ coordinates of the node position onto the node feature, which is indicated by +2 at the residual layer input. Residual layers and the detection head use the Look-up-Table Spline Convolutions (LUT-SC) as the basic building block (Fig.~\ref{fig:method_overview} B). These LUT-SC are trained as a standard Spline Convolution~\cite{Fey2018cvpr, Schaefer22cvpr} and later deployed as an efficient look-up-table (see Sec.~\ref{sec:asynchronous_operation}).\\
\textbf{Spline Convolutions} Spline Convolutions update the node features by aggregating messages from neighboring nodes:
\begin{align}
    \textbf{n}'^i_f = W\textbf{n}^i_f + \sum_{(j,i)\in E} W(e_{ij})\textbf{n}^j_f, \quad\text{ and }\quad\textbf{n}'^i_p = \textbf{n}^i_p
\end{align}
Here $\textbf{n}'^i_f$ is the updated feature at node $\textbf{n}_j$, $W\in\mathbb{R}^{c_\text{out}\times c_\text{in}}$ is a matrix that maps the current node feature $\textbf{n}^i_f$ to the output, and $W(e_{ij})\in\mathbb{R}^{c_\text{out}\times c_\text{in}}$ a matrix that maps neighboring node features $\textbf{n}^j_f$ to the output. In~\cite{Fey2018cvpr}, $W(e_{ij})$ is a matrix-valued smooth function of the edge feature $e_{ij}$. We set the edge feature to be $e_{ij}=\vert \textbf{n}^i_{xy}-\textbf{n}^j_{xy}\vert/2r+1/2$, where $\textbf{n}_{xy}$ denotes only the $x$ and $y$ component of the node position, and $r$ is chosen such that $e_{ij}\in[0,1]^2$. The function $W(e_{ij})$ is modeled by a $d$-order B-Spline in $m=2$ dimensions with $k\times k$ learnable weight matrices equally spaced in $[0,1]^2$. During the evaluation, the function interpolates between these learnable weights according to the value of $e_{ij}$. In this work, we choose $d=1$ and $k=5$. \\
\textbf{Max Pooling}
Max pooling splits the input space into $g_x \times g_y \times g_t$ voxels $V$, and clusters nodes in the same voxel. 
At the output, each non-empty voxel has a node, located at the rounded mean of the input node positions, and with its feature equal to the maximum of the input nodes features.    
\begin{align}
\label{eq:max_pooling}
    \textbf{n}'^i_f = \max_{\textbf{n}\in V_i} \textbf{n}_f \quad \text{ and } \quad \textbf{n}'^i_p = \frac{1}{\alpha}\left\lfloor\frac{\alpha}{|V_i|}\sum_{\textbf{n}\in V_i} \textbf{n}_p\right\rfloor
\end{align}
Here multiplying by $\alpha=[H,W,\frac{1}{\beta}]^T$ scales the mean to the original resolution. To compute the new edges, it forms a union of all edges connecting the cluster centers and removes duplicates. This operation can result in bi-directional edges between output nodes if at least one node from voxel A connected to one of voxel B and vice versa. The combination of max-pooling and position rounding has two main benefits: First, it allows the implementation of highly efficient LUT-SC convolutions, and second, it enables update pruning, which further reduces computation, discussed in Sec.~\ref{sec:ablation_studies}.
For our pooling layers we select $(g_x, g_y, g_t)_i=(56/2^i, 40/2^i, 1)$, where $i$ is the index of the pooling layer. As seen in Sec.~\ref{sec:ablation_studies}, selecting $g_t=1$ is crucial to obtain high performance, since it accelerates the information mixing in the network.\\
\textbf{Detection head} Inspired by the YOLOX detection head, we design a series of (LUT-SC, BN, ReLU) blocks which progressively compute a bounding box regression $\textbf{f}_\text{reg}\in\mathbb{R}^4$, class score $\textbf{f}_\text{cls}\in\mathbb{R}^{n_\text{cls}}$ and object score $\textbf{f}_\text{obj}\in\mathbb{R}$ for each output node. We then decode the bounding box location as in~\cite{Ge21arxiv}, but relative to the voxel location in which the node resides. This results in a sparse set of output detections.
\begin{figure*}
    \centering
    \begin{tabular}{c}
    \includegraphics[width=0.9\linewidth]{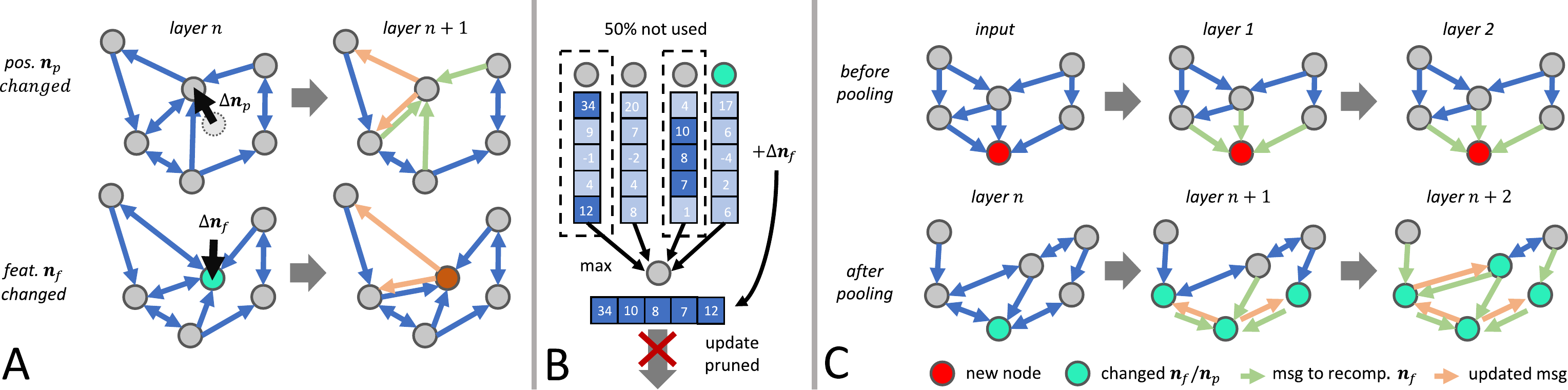}
    \end{tabular}
    \vspace{-1ex}
    \caption{Overview of update propagation rules for a single new event. For convolution layers (A), we update the messages sent after a node position of feature change. If either of them change, we send update messages from the changed node (orange arrows). When the position changes, recompute messages are also sent to the changed node (green messages). In pooling layers (B), output features originating from changed input nodes are recomputed. If the changed node is in the currently unused (grayed out) set, it does not have a feature higher than the current output and it does not change the output node sufficiently to change rounding, the update is pruned. C shows the application to multiple layers. Before pooling, edges are directed, so the number of computed messages remains constant with network depth. After pooling, bidirectional edges may appear, leading to a growth in the number of computed messages in lower layers.}\vspace{-2ex}
    \label{fig:update_propagation}
\end{figure*}

\subsection{Asynchronous Operation}
\label{sec:asynchronous_operation}
As in \cite{Schaefer22cvpr,Messikommer20eccv,Li21iccv}, after training, we deploy our network in an asynchronous mode. The conversion to asynchronous mode happens in three steps: 
(i) look-up-table spline convolution caching and batch norm fusing, (ii) network activation initialization, and (iii) update propagation and pruning. \\
\textbf{Look-up-Table Spline Convolution Caching}
Spline Convolutions generate the highest computational burden in our method since they involve evaluating a multi-variate, matrix-valued function, and performing a matrix-vector multiplication. Following the implementation in \cite{Fey2018cvpr}, computing a single message between neighbors requires 
\begin{equation}
\label{eq:flops_per_message}
    C_\text{msg}=(2[d+1]^m-1) c_\text{in}c_\text{out} + (2c_\text{in}-1)c_\text{out},
\end{equation}
floating point operations (FLOPS), where the first term computes the interpolation of the weight matrix, and the second computes the matrix-vector product. Here the first term dominates due to the highly superlinear dependence on $d$ and $m$. Our LUT-SC eliminates this term. We recognize that the edge attributes $e_{ij}$, only depend on the relative \emph{spatial} node positions. Since events are triggered on a grid, and the distance between neighbors is bounded, these edge attributes can only take on a \emph{finite} number of possible values. Therefore, instead of recomputing the interpolated weight at each step, we can precompute all weight matrices once and store them in a look-up table. This table stores the relative offsets of nodes together with their weight matrix. We thus replace the message propagation equation with 
\begin{align}
    \label{eq:convolution}
    \textbf{n}'^i_f &= W\textbf{n}^i_f + \sum_{(j,i)\in E} W_{ij}\textbf{n}^j_f\\
    W_{ij} &= LUT(dx,dy),
\end{align}
where $dx$ and $dy$ are the relative  2D positions of nodes $i$ and $j$. Note that this transformation reduces the complexity of our convolution operation to $C_\text{msg}=(2c_\text{in}-1)c_\text{out}$ which is on the level of the classical graph convolution (GC) used in \cite{Li21iccv}. However, crucially, LUT-SC still retains the relative spatial awareness of Spline Convolutions, since $W_{ij}$ changes with the relative position and is thus more expressive than GCs. After caching, we fuse the weights computed above with the batch norm layer immediately following each convolution, thereby eliminating its computation from the tally. After pooling, ordinarily, node positions would not have the property that they lie on a grid anymore, as their coordinates get set to the centroid location. However, since we apply position rounding, we can apply LUT-SC caching in all layers of the network.\\
\textbf{Network Activation Initialization} 
Before asynchronous processing, we pass a dense graph through our network and cache the intermediate activations at each layer. While in convolution layers we cache the activation, i.e., the results of sums computed from Eq.  \eqref{eq:convolution}, in max pooling layers we cache (i) the indices of input nodes used to compute the output feature for each voxel, (ii) a list of currently occupied output voxels, and (iii) a partial sum of node positions and node counts per voxel, to efficiently update output node positions after pooling. 

\textbf{Update Propagation and Pruning}
When a new event is inserted, we recompute messages in all layers of the network. We do this to achieve an output identical to the output the network would have computed if the complete graph with one event added was processed from scratch. The propagation rules are outlined in Fig.~\ref{fig:update_propagation}.

\emph{Input layer:} In the early layers of the network before pooling (C, top row), we only need to compute messages (green arrow) which are required to compute the feature of the new node in the network. Moreover, as the network depth increases, the number of messages stays constant, allowing us to stack multiple layers at the input with minimal computation increase.  
After pooling (C bottom row), bidirectional edges may be encountered, and thus update propagation follows the rules outlined in A and B.  This is because, it may happen that for voxels $V_i$ and $V_j$ one edge goes from  $V_i$ and $V_j$ and vice versa, with both edges being temporally ordered. The union operation of max pooling would then form a bidirectional edge between the output clusters.
At each layer, we maintain a running list of unchanged nodes (gray) and changed nodes (cyan), and whether their position has changed, the feature has changed, or both. 

\emph{Convolution Layers:} In a convolution layer (A), if the node has a different position (top), we recompute that node's feature and resend a message from that node to all its neighbors. These are marked as green and orange arrows in the top row of Fig.~\ref{fig:update_propagation} (c). If instead only the node's feature changed (bottom), we only update the messages sent from that node to its neighbors. We can gain an intuition for these rules from Eq.~\ref{eq:convolution}. A change in the node feature $\textbf{n}_f$ only changes one term in the sum which has to be recomputed. Instead, a node position change causes all weight matrices $W_{ij}$ to change, resulting in a recomputation of the entire sum. We provide more details about these rules in the appendix. 

\emph{Pooling Layers and Pruning:} During pooling, update pruning can occur. When an input node has a changed position or feature we check if (i) the changed node is currently in the set of unused nodes (grayed out in Fig.~\ref{fig:update_propagation}), (ii) the changed feature of the node does not beat the current maximum at any feature index, and (iii) its position change did not deflect the average output node position sufficiently to change rounding. If not all three conditions are met, we recompute the output feature for that node, otherwise, we prune the update and skip the computation in the lower layers. Skipping happens surprisingly often. In our case, we found that 73\% of updates are skipped due to this mechanism. This also motivated us to place the max pooling layer in the early layers, since then it has the highest potential to save computation. In the next section, we will show the impact these features have on the computation of the method. 

\begin{figure*}[t]
\centering
\setlength{\tabcolsep}{4pt} 
\begin{tabular}{cccc}
    \includegraphics[height=0.165\linewidth]{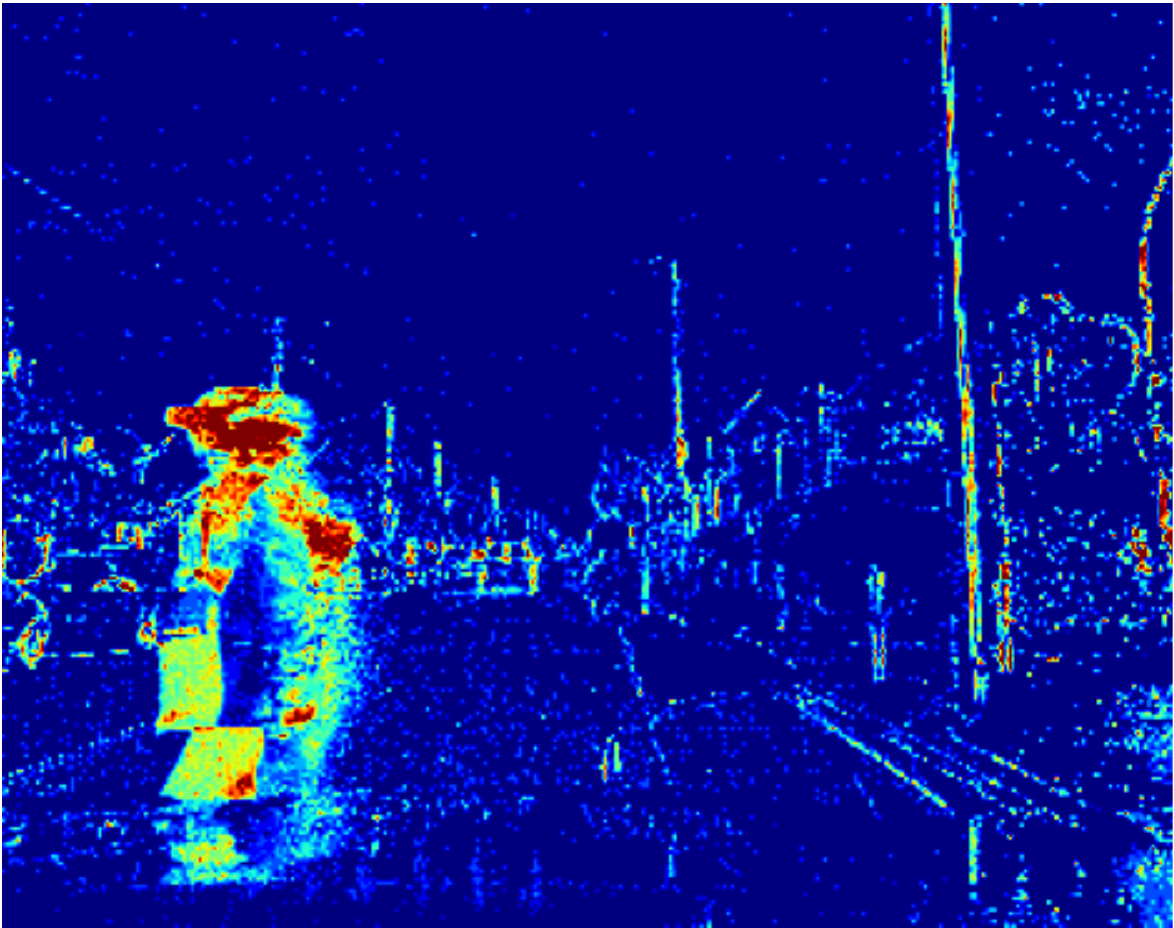}&
    \includegraphics[height=0.165\linewidth]{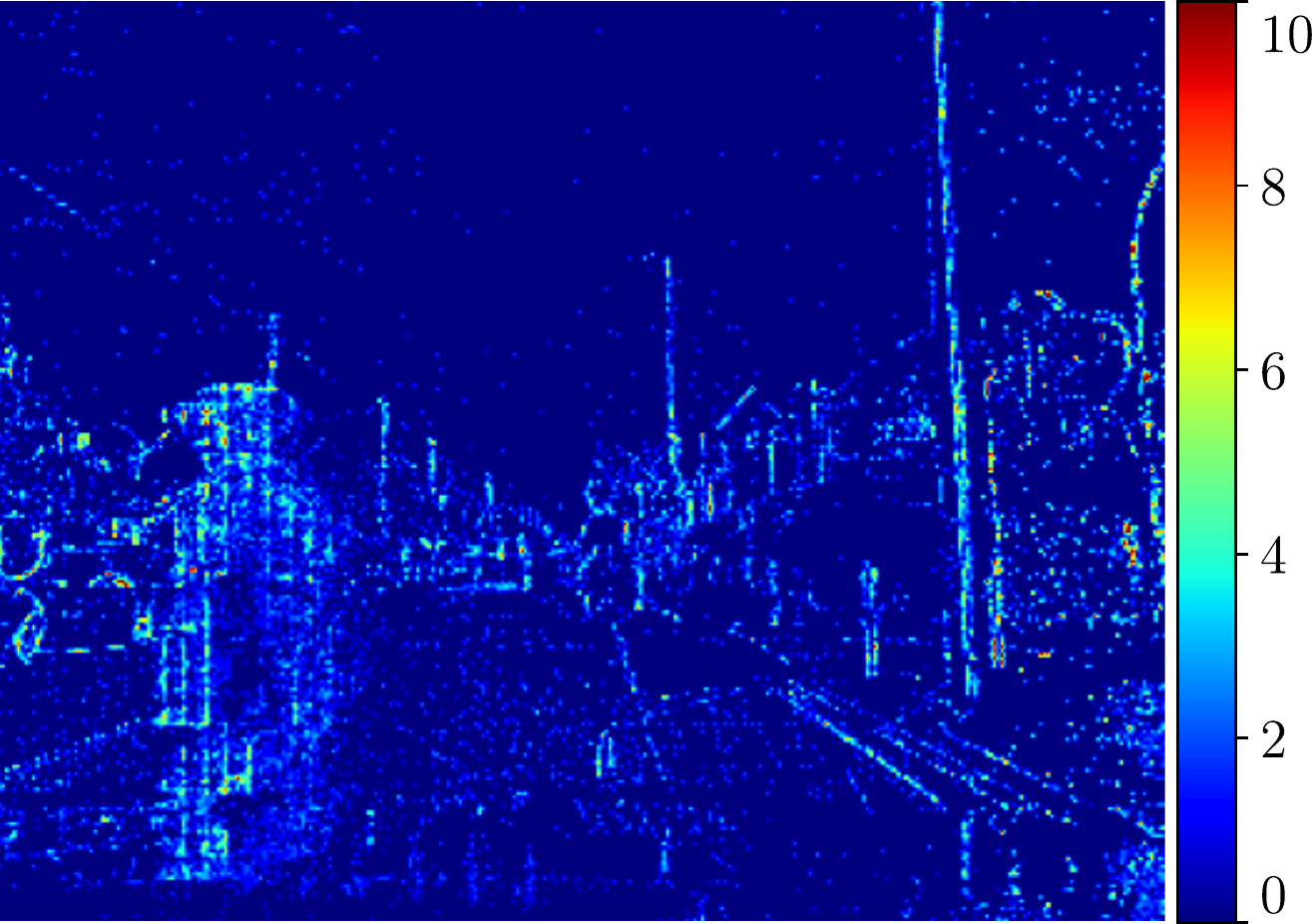}&
    \includegraphics[height=0.165\linewidth]{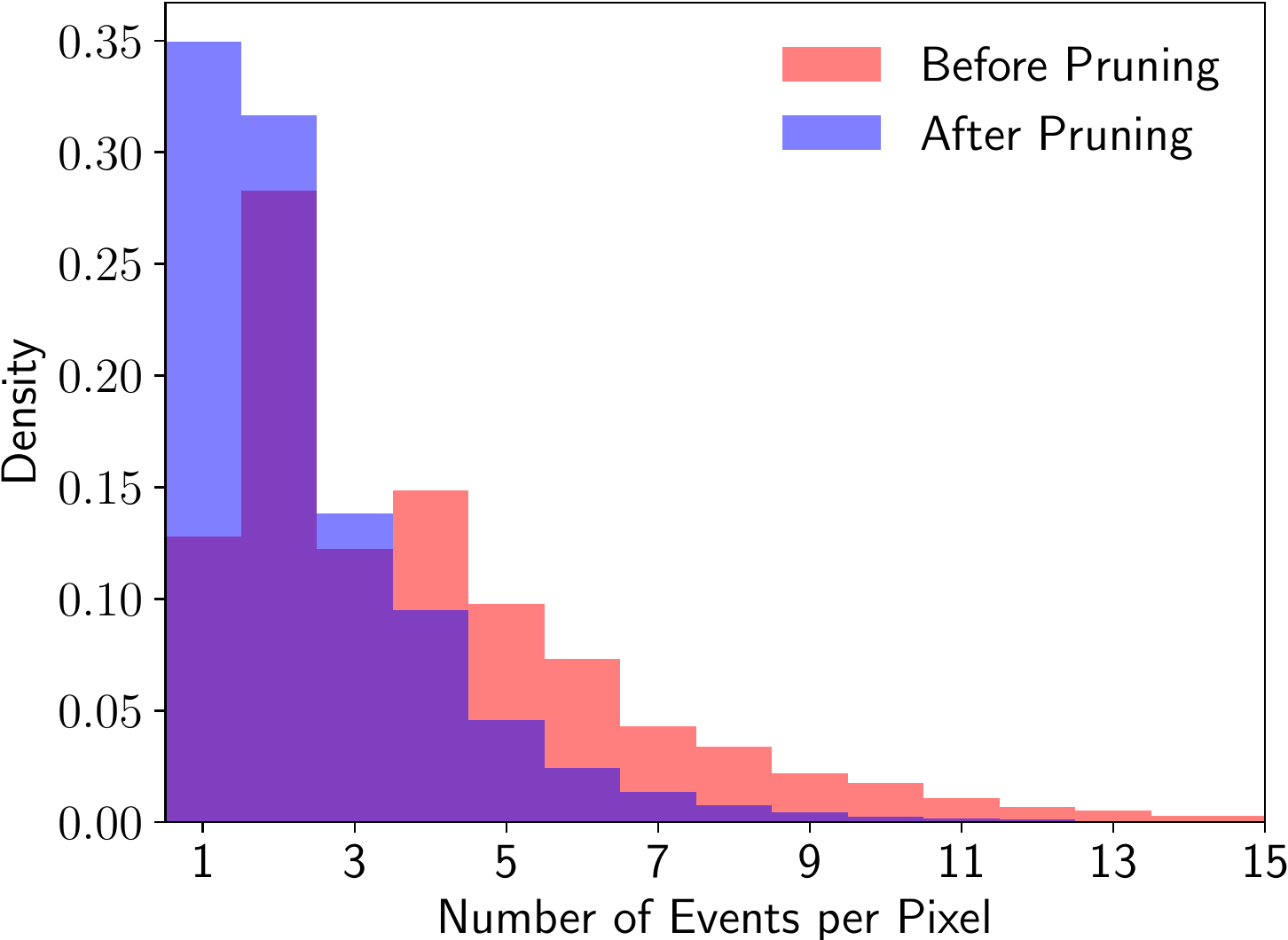}&
    \includegraphics[height=0.165\linewidth]{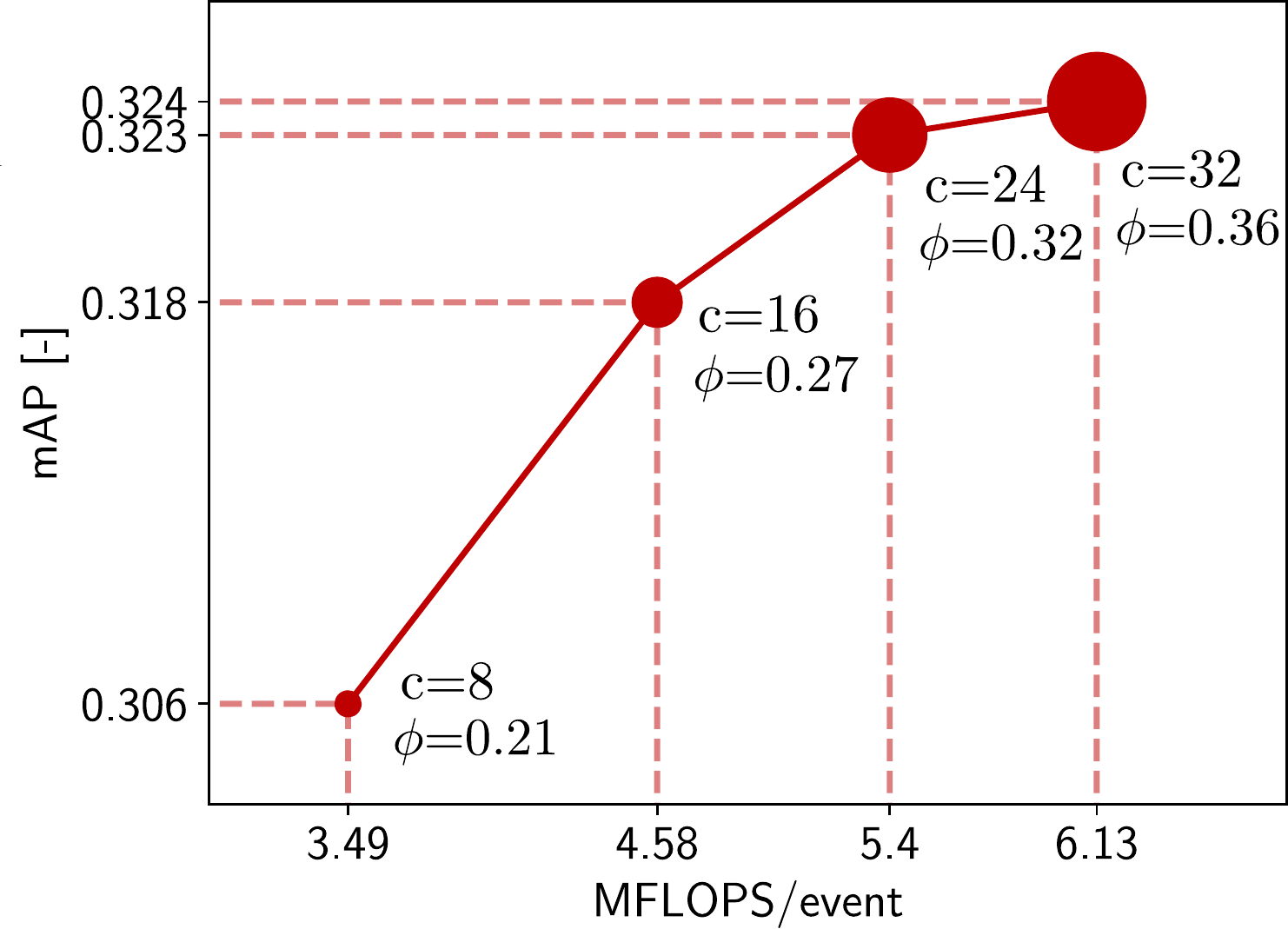}\\
    (a) events without pruning &(b) events after pruning  &(c) num. events per pixel & (d) effect of num. channels 
\end{tabular}
\vspace{-1.5ex}
    \caption{Effect of update pruning due to max pooling. 
    We interpret max pooling as a kind of event filter. In (a-b) we show an example of aggregated events before (a) and after (b) filtering. This filter acts as a saliency detector, only letting through events with "new information", and removing redundant events in high event rate regions. This results in a more uniform distribution of events (c). 
    We can control the filter strength by modulating the number of output features, $c$. As seen in (d), increasing $c$ increases both computation and mAP. However, mAP growth drastically reduces in slope after $c=24$. The dot size is proportional to $c$, and $\phi$ measures the proportion of updates that pass through the filter. In our baseline setting with $c=16$, we see that only 27\% of updates pass the first max pooling layer.}
    \vspace*{-3ex}
    \label{fig:max_pooling}
\end{figure*}
\begin{table}[]
\resizebox{\linewidth}{!}{%
\begin{tabular}{cccc|cc}
\hline
\textbf{2D Conv} & \textbf{LUT-SC} & \textbf{Pruning} & \textbf{Pos. Rounding} & \textbf{mAP$\uparrow$} & \textbf{MFLOPS/ev$\downarrow$} \\ \hline
\xmark           & \xmark          & \xmark           & \xmark                     & 31.84        & 150.87                 \\
\cmark           & \xmark          & \xmark           & \xmark                     & 31.90        & 79.6                 \\
\cmark           & \cmark          & \xmark           & \xmark                     & 31.90        & 17.3               \\
\cmark           & \cmark          & \cmark           & \xmark                     & \textbf{31.90}        & 16.3               \\
\cmark           & \cmark          & \cmark           & \cmark                     & 31.79        & \textbf{4.58}      \\ \hline
\end{tabular}}
\vspace{-1ex}
\caption{Features affecting computational complexity.}\label{tab:ablations_efficiency}
\end{table}

\begin{table}[]
    \centering
    \resizebox{0.8\linewidth}{!}{
\begin{tabular}{l|cc}
\hline
                                          \textbf{Ablation}     & \textbf{mAP$\uparrow$}   & \textbf{MFLOPS/ev$\downarrow$} \\ \hline
 w/o early aggregation & 21.2          & 6.27               \\
w/o multi-layer input & 30.0          & \textbf{4.56}\\
\textbf{baseline}                     & \textbf{31.8} & 4.58               \\ \hline
\end{tabular}}  
\vspace{-1ex}
\caption{Features affecting accuracy of the method.}
\vspace{-3ex}
    \label{tab:ablations_accuracy}
\end{table}

\section{Experiments}

\textbf{Training Details}
In all our experiments, we use the AdamW optimizer~\cite{Loshchilov19iclr} with a learning rate of 0.01 and weight decay of $10^{-5}$. We train each model for 150'000 iterations with a batch size of 64. We randomly crop the events to 75\% of the full resolution, and randomly translate them by up to 10\% of the full resolution. We use the YOLOX loss~\cite{Ge21arxiv}, which includes an IOU loss, class loss, and a regression loss, discussed in~\cite{Ge21arxiv}. To stabilize training, we also use exponential model averaging (EMA)~\cite{Izmailov18uai}.

\textbf{Datasets} We evaluate our method on the N-Caltech101 detection~\cite{Orchard15fns}, and the Gen1 Detection Dataset~\cite{Tournemire20arxiv}. N-Caltech101 consists of recording by a DAVIS240~\cite{Brandli14ssc} undergoing a saccadic motion in front of a projector, projecting samples of Caltech101~\cite{FeiFei04cvprw} on a wall. In post-processing, bounding boxes around the visible boxes were hand placed. 
The Gen1 Driving Dataset is a more challenging, large-scale detection dataset targeting an automotive setting. It was recorded with an ATIS sensor~\cite{Posch11ssc} with a resolution of $304\times 240$, two classes, 228,123 annotated cars, and 27,658 annotated pedestrians. As \cite{Perot20nips} we remove bounding boxes diagonals below 30 and sides below 20 pixels from Gen1.

\subsection{Ablation Studies}
\label{sec:ablation_studies}
Here we motivate the use of the features discussed previously. We split our ablation studies into two parts, i.e. those targeting the efficiency (Tab.~\ref{tab:ablations_efficiency}), or the accuracy (Tab.~\ref{tab:ablations_accuracy}) of the method. For all experiments, we use the model illustrated in Fig.~\ref{fig:method_overview} as a baseline, and report means average precision (mAP) scores~\cite{Lin14eccv} on the validation set of Gen1~\cite{Tournemire20arxiv}. 
 
\textbf{Ablations on Efficiency}
A key enabling factor for using 2D LUT-SCs lies in transitioning from 2D to 3D convolutions, which we investigate by training a model with 3D Spline Convolutions (row 1 in Tab.~\ref{tab:ablations_efficiency}). This result does not yet take into account update pruning, which is discussed later. With an mAP of 31.84, it achieves a 0.05 higher mAP than our baseline (bottom row). Using 3D convolutions yields a slight improvement in accuracy since it uses more information, but does not allow us to perform an efficient lookup, yielding 150.87 MFLOPS per new event. Using 2D convolutions (row 2) reduces the computation to 79.6 MFLOPS/ev due to the dependence on the dimension $d$ in Eq.~\eqref{eq:flops_per_message}, which is further reduced to 17.3 MFLOPS/ev after implementing LUT-SCs (row 3). Despite the small decrease in performance due to 2D convolutions, we gain a factor of 8.7 in terms of FLOPS per event.

Next, we investigate pruning. We recompute the FLOPS of the previous model by terminating update propagation after max-pooling layers, illustrated in Fig.~\ref{fig:update_propagation} (c), and reported in row 4 of Tab.~\ref{tab:ablations_efficiency}. We find that this reduces the computational complexity from 17.3 to 16.3 MFLOPS/ev. This reduction comes from removing the orange messages in Fig.~\ref{fig:update_propagation} A (bottom). Implementing node position in Eq.~\ref{eq:max_pooling} (row 5), allows us to fully prune updates. The final method only requires 4.56 MFLOPS/ev. Node position rounding reduces mAP only by 0.01, justifying its use.\\

\textbf{Ablations on Accuracy}
We found that two features of our network had a major impact on performance: First, we applied early temporal aggregation, i.e., using $g_t=1$, which sped up training and led to higher accuracy. We train another model which pools the temporal dimension more gradually by setting  $g_t=8/2^i$, where $i$ is the index of the pooling layer. This model only reached an mAP of 21.2 (Tab.~\ref{tab:ablations_accuracy}, first row), after reducing the learning rate to 0.002 to enable stable training. This highlights that early pooling plays an important role since it improves our result by 10.6 mAP. We believe that it is important for mixing features quickly so that they can be used in lower layers. 

Next, we investigate using multiple layers before the max-pooling layer. We train another model which only has a single input layer, replacing the Layer in Fig.~\ref{fig:method_overview} with a (LUT-SC, BN, ReLU) block. This yielded a performance of 30.0 mAP which is 1.8 mAP lower than the baseline. The computational complexity is only marginally lower, which is explained by Fig.~\ref{fig:update_propagation} C (top). We see that adding layers at the input only generates few additional messages. This highlights the benefits of using a directed event graph.
\subsection{Max Pooling}\vspace{-1ex}
In this section, we take a closer look at the pruning mechanism. We find that almost all pruning happens in the very first max pooling layer. This motivates the placement of the pooling layer at the early stages of the network, which allows us to skip most computations when pruning happens. Also, since the subgraph is still small in the early layers, it is easy to prune the entire update tree. We interpret this case as "event filtering" and investigate this filter in Fig.~\ref{fig:max_pooling}. 

When applied to raw events (Fig. (a)) we obtain filtered events (Fig. (b)), i.e., events that passed through the first max pooling layer. We observe that max-pooling makes the events more uniformly distributed over the image plane. This is also supported by the density plot in Fig.~\ref{fig:max_pooling} (c), which shows that the distribution of the number of events per pixel shifts to the left after filtering, removing events in regions where there are too many. This behavior can be explained by the pigeon-hole principle when applied to max-pooling layers. 
Max-pooling usually only uses a fraction of its input nodes to compute the output feature. The number of input nodes used by the max-pooling layer is upper bounded by its output channel dimension, $c_\text{out}$, since it could at maximum only use one feature from each input node. As a result, max-pooling selects at most $c_\text{out}$ nodes for each voxel, resulting in more uniformly sampled events. 

To study the effect of the output channel dimension on filtering, we train four models with $c_\text{out}\in \{8, 16, 24, 32\}$, where our baseline model had $c_\text{out}=16$. We report the mAP, MFLOPS/ev, and fraction of events after filtering, $\phi$ averaged over Gen1, in Fig.~\ref{fig:max_pooling} (d). As predicted, we find that increasing $c_\text{out}$, increases mAP, MFLOPS, and $\phi$. However, increase happens at different rates. While MFLOPS and $\phi$ grow roughly linearly, mAP growth slows down signficantly after $c=24$. Interestingly, by selecting $c_\text{out}=8$ we still achieve an mAP of 30.6, while only using  $21\%$ of events.     
This type of filtering has interesting implications for future work. An interesting question would be whether events that are not pruned carry salient and interpretable information. 

\subsection{Comparison with State of the Art}\vspace{-1ex}
\label{sec:sota_comparison}
Finally, we compare our method against state-of-the-art dense and asynchronous methods and report results in Tab.~\ref{tab:sota_comparison} on the N-Caltech101 and Gen1 test sets. We evaluate four versions of our model: nano (-N), small (-S), medium (-M), and large (-L). These differ in the number of features in the layer blocks 3,4 and 5 and in the detection heads, and have 32, 64, 92, and 128 channels in these layers respectively. Our baseline from before has 64. We compare against the following state-of-the-art methods:\\
\textbf{Dense Recurrent Methods} In this category, RED~\cite{Perot20nips} and ASTM~\cite{Li22trip} are the state-of-the-art, and feature recurrent architectures. We also include MatrixLSTM+YOLOv3~\cite{Cannici20eccv} which features a recurrent, learnable representation and a YOLOv3 detection head.\\
\textbf{Dense Feed-forward Methods} The work in \cite{Li22trip} provides results on Gen1 for the dense feed-forward methods which we term Events+RRC~\cite{Chen18cvprw}, Inception+SDD~\cite{Iacono18iros} and Events+YOLOv3~\cite{Jiang19icra}. These use dense event representations with the RRC, SSD, or YOLOv3 detection head. \\
\textbf{Spiking Methods} We compare against the spiking network Spiking DenseNet~\cite{Cordone22ijcnn}, which uses an SSD detection head. \\
\textbf{Asynchronous Methods} Here we compare against state-of-the-art methods AEGNN~\cite{Schaefer22cvpr} and NVS-S~\cite{Li21iccv}, both graph-based, AsyNet~\cite{Messikommer20eccv} which uses submanifold sparse convolutions~\cite{Graham18cvpr} and YOLE~\cite{Cannici19cvprw}, which uses an asynchronous CNN. All of these methods deploy their networks in an asynchronous mode during testing.\\
\begin{table}[]
\setlength{\tabcolsep}{2pt}
\resizebox{\linewidth}{!}{
\begin{tabular}{lc|cc|cc}
\hline
\multirow{2}{*}{\textbf{Method}}          & \multirow{2}{*}{\textbf{Async.}} & \multicolumn{2}{c|}{\textbf{Gen1}}     & \multicolumn{2}{c}{\textbf{N-Caltech101}}               \\
                                                               &                                  & \textbf{mAP$\uparrow$} & \textbf{MFLOPS/ev$\downarrow$}  & \textbf{mAP$\uparrow$} & \textbf{MFLOPS/ev$\downarrow$}  \\ \hline
Inception+SSD   \cite{Iacono18iros}      & \xmark & 30.1 & $>$8'245* & -     & -   \\
Events+RRC       \cite{Chen18cvprw}       & \xmark & 30.7 & $>$21’758 & -     & -   \\
MatrixLSTM+YOLOv3      \cite{Cannici20eccv}     & \xmark & 31.0 & $>$34'519*  & -     & -   \\
Events+YOLOv3    \cite{Jiang19icra}       & \xmark & 31.2 & $>$34'518*        & -     & -   \\
RED             \cite{Perot20nips}       & \xmark & 40.0 & 4'712     & -     & -   \\
ASTM-Net        \cite{Li22trip}          & \xmark & 46.7 & $>$21'758* & -     & -   \\
NVS-S           \cite{Li21iccv}          & \cmark & 8.60 & 7.80     & 34.6 & 7.80\\
AsyNet          \cite{Messikommer20eccv} & \cmark & 14.5 & 205      & 64.3 & 200 \\
AEGNN           \cite{Schaefer22cvpr}    & \cmark & 16.3 & 5.26     & 59.5 & 7.41\\
Spiking DenseNet\cite{Cordone22ijcnn}    & \cmark & 18.9 & N/A      & -     & -   \\
YOLE            \cite{Cannici19cvprw}    & \cmark & -     & -        & 39.8 & 3682\\ \hline
\multicolumn{1}{l}{\textbf{EAGR-N (ours)}}&\cmark& 26.3& \textbf{1.36} &  62.9 & \textbf{2.28}\\
\multicolumn{1}{l}{\textbf{EAGR}}&\cmark& 30.4& 4.58 & 70.2 & 6.85\\
\multicolumn{1}{l}{\textbf{EAGR-M (ours)}}&\cmark& 31.8& 9.94 & 72.7 & 12.2 \\
\multicolumn{1}{l}{\textbf{EAGR-L (ours)}}&\cmark& \textbf{32.1}& 17.4 & \textbf{73.2} & 18.9\\\hline
\multicolumn{6}{l}{* \small{lower bound from network backbone}}\\
 \multicolumn{6}{l}{\small{N/A: FLOPS are undefined due to spike-based computation.}}\\

\hline
\end{tabular}}\vspace{-1ex}
\caption{Comparison against state of the art methods on the Gen1 detection dataset~\cite{Tournemire20arxiv} and N-Caltech101~\cite{Orchard15fns}.}\label{tab:sota_comparison}
\vspace{-3.5ex}
\end{table}

Since implementation details are not available for Events+RRC~\cite{Chen18cvprw}, Inception+SDD~\cite{Iacono18iros} and Events+YOLOv3~\cite{Jiang19icra}, MatrixLSTM+YOLOv3~\cite{Cannici20eccv} and ASTM-Net~\cite{Li22trip} we find a lower bound on the per-event computation necessary to update their network based on the complexity of their detection backbone. While for Events+YOLOv3, and MatrixLSTM+YOLOv3 we use the DarkNet-53 backbone, for ASTM-Net and Events+RRC we use the VGG11 backbone, and for Inception+SDD the Inception v2 backbone. Since Spiking DenseNet uses spike-based computation, we do not report FLOPS since they are undefined and mark that entry with N/A.  

From Tab.~\ref{tab:sota_comparison} we can make multiple observations. First, we find that recurrent dense methods RED and ASTM net outperform our L model by 7.9 mAP and 14.6 mAP respectively. However, since these methods are dense, we observe significantly higher computation when compared to our method (4712 vs. 1.36 for our nano model). We believe that deeper networks and recurrence are two major factors that help performance in their methods. By contrast, our large model with 32.1 mAP outperforms the recurrent method MatrixLSTM~\cite{Cannici20eccv} by 1.1 mAP and also has 120 times fewer FLOPS. We also find that our large model with an mAP of 32.1 outperforms dense feed-forward methods Events+RRC~\cite{Chen18cvprw} (30.7), Inception+SSD~\cite{Iacono18iros} (30.1) and Events+YOLOv3~\cite{Jiang19icra} (31.2). When compared to the spiking network~\cite{Cordone22ijcnn} we find that our method has a 13.1 mAP higher score. The low performance of the SNN is expected to increase as better learning strategies become available to the community. Finally, we compare against sparse methods. We find that our small model outperforms all methods in terms of computation, with around 13\% times fewer MFLOPS/ev than the runner-up AEGNN~\cite{Schaefer22cvpr}. It also achieves a 14.1 mAP higher performance than AEGNN. Our smallest network, nano,  is even 3.8 times more efficient while still outperforming AEGNN by 10 mAP.

On the N-Caltech101 dataset, our small model outperforms state-of-the-art dense and sparse methods, achieving 70.2 mAp, which is 5.9 mAP higher than the runner-up AsyNet~\cite{Messikommer20eccv}. Moreover, this model has lower computational complexity than state-of-the-art AEGNN~\cite{Schaefer22cvpr}. Our large model achieves the highest score with 73.2 mAP. Our nano model achieves the lowest computation of 2.28 MFLOPS/ev, which is 3.25 times lower than that of AEGNN and has a 3.4 higher mAP. It is higher than that for Gen1 since N-Caltech101 has more classes.\\
\textbf{Timing Experiments}
We compare the time it takes for our dense GNN to process a batch of 50'000 events averaged over Gen1, and compare it against our asynchronous implementation on a Quadro RTX 4000 laptop GPU. We found that our dense network takes 30.8 ms, while the asynchronous method requires 8.46 ms, a 3.7-fold reduction. We believe that with further optimizations, and when deployed on potentially spiking hardware, this method can reduce power and latency by additional factors.
\vspace{-1ex}
\section{Limitations and Future Work} \vspace{-1ex}
While making an important step toward highly efficient and robust event processing, we believe that further improvements can be made. In particular, we found that the methods currently outperforming our event-based detector are all recurrent, a feature that was not studied in this work. We thus believe that our method could also benefit from recurrency since it can overcome issues especially when few events are present. Combining this approach with additional sensors, such as LiDARs or RGB cameras, is also a promising research direction. They can provide strong priors, which may increase its performance and reduce its complexity if shallower networks can be utilized. 
\vspace{-1ex}
\section{Conclusion}\vspace{-1ex}
While traditionally, data-driven event-based methods have relied on converting events into dense image-like representations, these still do not effectively model the asynchronous and sparse nature of event data, which increases computation. Despite recent progress on reducing this complexity, asynchronous methods have not yet delivered on high accuracy due to their shallow networks. In this work, we introduced a series of network design choices that allow us to design deeper neural networks without sacrificing complexity. We achieve this by introducing novel look-up-based convolutions, performing update pruning, and using directed event graphs in early layers which reduce the complexity of our approach by a factor of 33.1 compared to the baseline. Thus, on the Gen1 and N-Caltech101 datasets, our method achieves a 11.5 point higher mAP than state-of-the-art asynchronous methods, with higher efficiency, and it even outperforms several dense, feed-forward methods. This opens the door to efficient,
and accurate object detection in edge-case scenarios.

\section{Acknowledgement}
This work was supported by Huawei Zurich, the Swiss National Science Foundation through the National Centre of Competence in Research (NCCR) Robotics (grant number 51NF40\_185543), and the European Research Council (ERC) under grant agreement No. 864042 (AGILEFLIGHT).

\section{Appendix}
\subsection{Overview}
Here we will discuss in more detail the equivalence of the update rules discussed in Sec. M-3.3 (Sec.~A-\ref{sec:update_rule_equivalence}). Then we will analyze per layer statistics in more detail in Sec.~A-\ref{sec:per_layer}, such as per layer computation, node updates and filtering.
In what follows, we will mark sections, and equations from the main manuscript with a "M-", and sections and equations here with "A-".  

\subsection{Asynchronous Operation}
\label{sec:update_rule_equivalence}
While the work in \cite{Schaefer22cvpr} flagged changed nodes in each layer, and recomputed each one of them by collecting all messages from the surrounding nodes of the updated nodes, in this work we take a more efficient route. We start off by considering the Eq. M-6 from the main text, which we repeat here: 
\begin{align}
\label{eq:conv}
    \textbf{n}_f'^i &= W\textbf{n}_f^i+\sum_{(j,i)\in E}W_{ij}\textbf{n}_f^j\\
    W_{ij}&=LUT(dx,dy)
\end{align}
Here, $dx$ and $dy$ are 
\begin{equation}
    \label{eq:edge_attr}
    \left[\begin{matrix}
    dx \\ dy
    \end{matrix}\right]=\left[\begin{matrix}
    \frac{\textbf{n}^j_{p,x} - \textbf{n}^i_{p,x}}{2r}+\frac{1}{2} \\
    \frac{\textbf{n}^j_{p,y} - \textbf{n}^i_{p,y}}{2r}+\frac{1}{2}
    \end{matrix}\right]
\end{equation}
and only take on a \emph{finite} number of values. In the following we discuss what messages need to be recomputed if the position and the feature of the node change. 

\begin{figure}
    \centering
    \includegraphics[width=\linewidth]{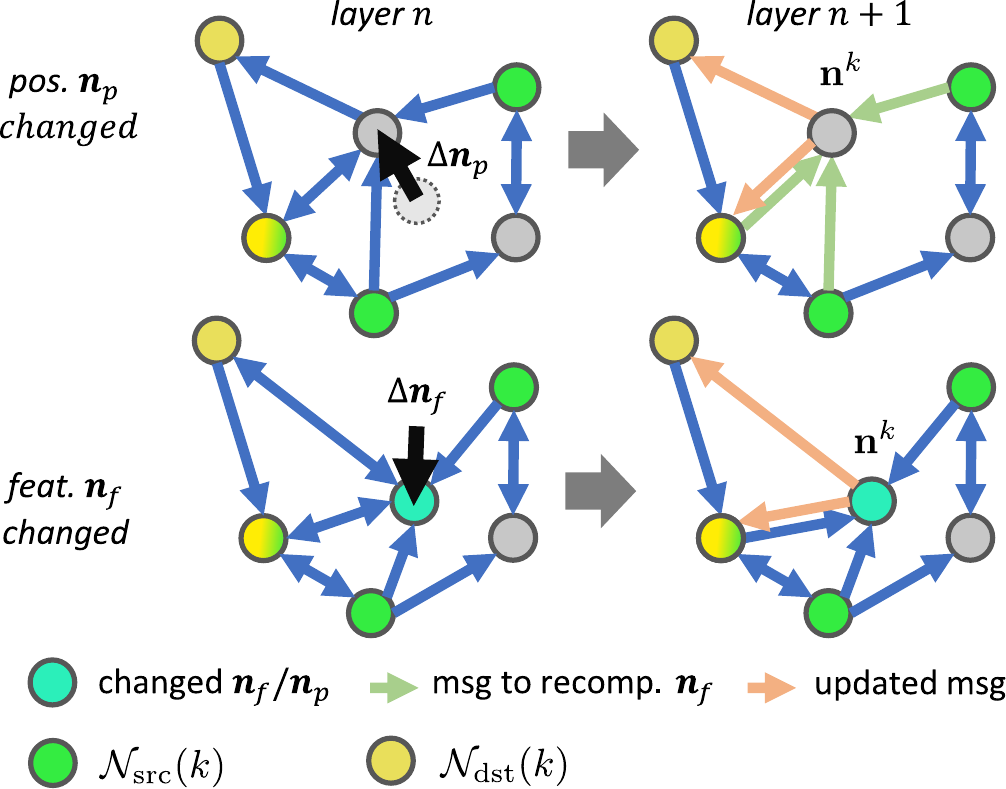}
    \caption{Details about update mechanism. We analyze the update computation necessary when a node $\textbf{n}_k$ is updated, and differentiate between position and feature updates. Green arrows denote updated messages from the green source neighbors of $\textbf{n}_k$, denoted as $\mathcal{N}(k)_\text{src}$. Orange arrows denote update messages from node $\textbf{n}_k$ to its yellow destination arrows, $\mathcal{N}(k)_\text{dst}$. Nodes can belong to both sets, and these are illustrated with a yellow-green gradient.}
    \label{fig:update_propagation_detailed}
\end{figure}

\subsubsection{Position Change}
Assume the position of a node $\textbf{n}^k$ changes, then we essentially update the weight matrices $W_{ij}$ since the $dx$ and $dy$ change in Eq.~A-\eqref{eq:edge_attr}. We thus need to modify weight matrices of the form $W_{kj}$ and $W_{ik}$. We will indicate the update weights with $W^*$. Since edges $(j,i)$ are used in Eq.~A-\eqref{eq:conv}, we conclude that only nodes with edges $(j,k)$ and $(k,i)$  are affected. While the edges $(j,k)$ correspond to messages sent from \emph{source neighbors} $\textbf{n}^j$ to node $\textbf{n}^k$, the edges $(k,i)$ correspond to messages sent from $\textbf{n}^k$ to its \emph{destination neighbors} $\textbf{n}^i$.
For this reason, only node $\textbf{n}^k$ and its \emph{destination neighbors}
\begin{equation}
    \mathcal{N}_\text{dst}(k)=\{\textbf{n}^i | (k,i)\in E\}
\end{equation}
need to be updated.
We discuss the update to each of them separately: \\
\textbf{Changed node $\textbf{n}^k$: } Eq.~A-\eqref{eq:conv} for $\textbf{n}^k$ becomes 
\begin{equation}
    \textbf{n}_f'^k = W\textbf{n}_f^k+\underbrace{\sum_{(j,k)\in E}W^*_{kj}\textbf{n}_f^j}_{\text{needs to be recomputed}}
\end{equation}
where we need to recompute all messages in the sum. The number of summands is the number of the \emph{source neighbors} of $\textbf{n}^k$
\begin{equation}
    \mathcal{N}_\text{src}(k)=\{\textbf{n}^j | (j,k)\in E\}
\end{equation}
which we define as $N_{k,\text{src}}\doteq\vert \mathcal{N}_\text{src}(k) \vert$
so the amount of computation to perform the above step is 
\begin{align}
    C_{\text{recomp. } k} &= N_{k,\text{src}} ( 2c_\text{in}-1)c_\text{out} + (N_{k,\text{src}} - 1)c_\text{out} \\
    \nonumber&=( 2c_\text{in}N_{k,\text{src}}-1)c_\text{out}.
\end{align}
Here the first term corresponds to performing the matrix-vector multiplications, and the second to summing the outputs. These exactly correspond to the messages illustrated as green arrows in Fig. A-\ref{fig:update_propagation_detailed} \\
\textbf{Update messages to destination nodes: } Eq.~A-\eqref{eq:conv} for nodes $\textbf{n}^i\in\mathcal{N}_\text{dst}(k)$ becomes 
\begin{equation}
\label{eq:dst_update}
    \textbf{n}_f'^i = W\textbf{n}_f^i+\sum_{(j,i)\in E}W_{ij}\textbf{n}_f^j +\underbrace{W^*_{ik}\textbf{n}_f^k - W_{ik}\textbf{n}_f^k}_{\text{needs to be recomputed}}
\end{equation}
Here we replace the old message by subtracting off the old message and adding on the new one. In total this results in a computation
\begin{align}
    C_{\text{update pos. } k} &= N_{k,\text{dst}} ( 2c_\text{in}-1)c_\text{out} + 2N_{k,\text{dst}}c_\text{out} \\
    \nonumber&=N_{k,\text{dst}} ( 2c_\text{in}+1)c_\text{out}.
\end{align}
Here the first term computes the new message, a total of $N_{k,\text{dst}}\doteq\vert \mathcal{N}_\text{dst}(k) \vert$ times, and the second term computes the subtraction and addition for each output message. These exactly correspond to the messages illustrated as orange arrows in Fig. A-\ref{fig:update_propagation_detailed}.\\
\textbf{Total:} In total, the computation necessary for a single node position change is 
\begin{equation}
    C_{\text{pos. change } k} = C_{\text{recomp. } k} + C_{\text{update pos. } k}
\end{equation}
\subsubsection{Feature Change}
Assume the node feature of node $\textbf{n}^k$ changes, indicating the change of $\textbf{n}^k_f$. We will denote the changed feature by $\textbf{n}^{k,*}_f$. Inspecting Eq.~A-\eqref{eq:conv} we find that now only nodes $\textbf{n}_f^i$ with $(k,i)\in E$ are affected, which is what we previously called the destination neighbors $\mathcal{N}_\text{dst}(k)$. Additionally, we need to update the root linear term $W\textbf{n}^k_f$, which simply amounts to a computation
\begin{align}
    C_\text{root} &=  (2c_\text{in}-1)c_\text{out} + 2c_\text{out}\\
    \nonumber &= (2c_\text{in}+1)c_\text{out}
\end{align}
The update message terms instead follow the updated Eq.~A-\eqref{eq:conv}
\begin{equation}
    \textbf{n}_f'^i = W\textbf{n}_f^i+\sum_{(j,i)\in E}W_{ij}\textbf{n}_f^j +\underbrace{W_{ik}\textbf{n}_f^{k,*} - W_{ik}\textbf{n}_f^k}_{\text{needs to be recomputed}}
\end{equation}
as in Eq.~A-\eqref{eq:dst_update}, this amounts to a computation 
\begin{align}
    C_{\text{update feat. } k} &= N_{k,\text{dst}} ( 2c_\text{in}-1)c_\text{out} + 2N_{k,\text{dst}}c_\text{out} \\
    \nonumber&=N_{k,\text{dst}} ( 2c_\text{in}+1)c_\text{out}.
\end{align}\\
\textbf{Total:} In total, the computation for updating the node feature is
\begin{equation}
    C_{\text{feat. change } k} = C_{\text{root}} + C_{\text{update feat. } k}
\end{equation}

\subsection{Per Layer Analysis}
\label{sec:per_layer}
In this section, we will have a deeper look at the per-layer metrics, of our network which will give us more insights into our method and also inspire future research in building better and more efficient neural network architectures. In what follows, we will analyze three quantities: (i) per-layer FLOPS (ii) per-layer changed nodes, and (iii) per-layer probability of filtering. We will discuss each in turn.
\begin{figure}[t!]
    \centering
    \includegraphics[width=1\linewidth]{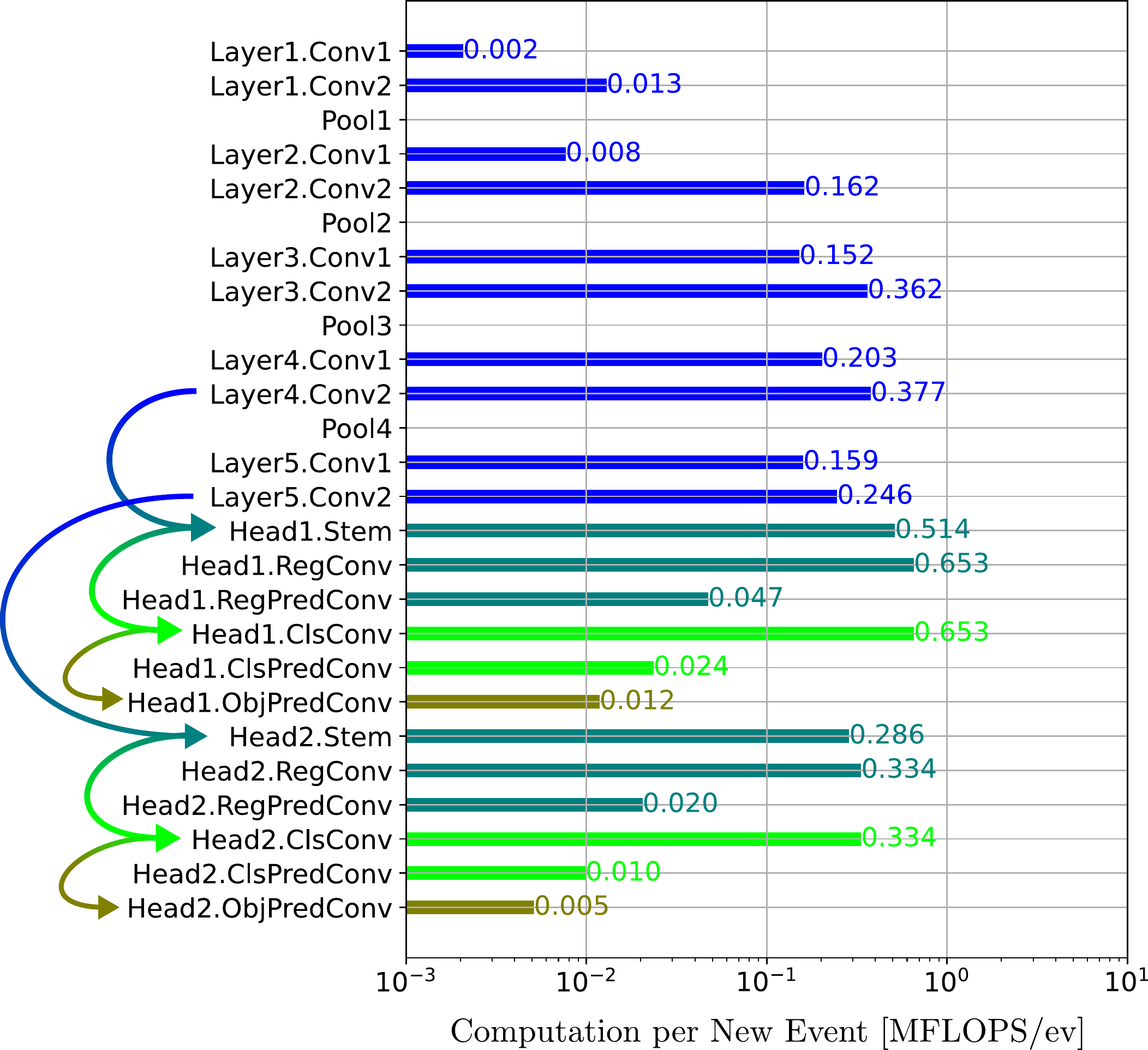}
    \caption{Computation for each new event broken down per layer of our neural network. The neural network layers follow the labelling convention of Fig. M-2 in the main text. We find the values for each layer by averaging the number of FLOPS over the Gen1 Dataset~\cite{Tournemire20arxiv}. Layers on the same network branch are have the same color, and arrows indicate branch connections. }
    \label{fig:flops_per_layer}
\end{figure}

\begin{figure*}[t!]
    \centering
    \begin{tabular}{c}
         \includegraphics[width=0.7\linewidth]{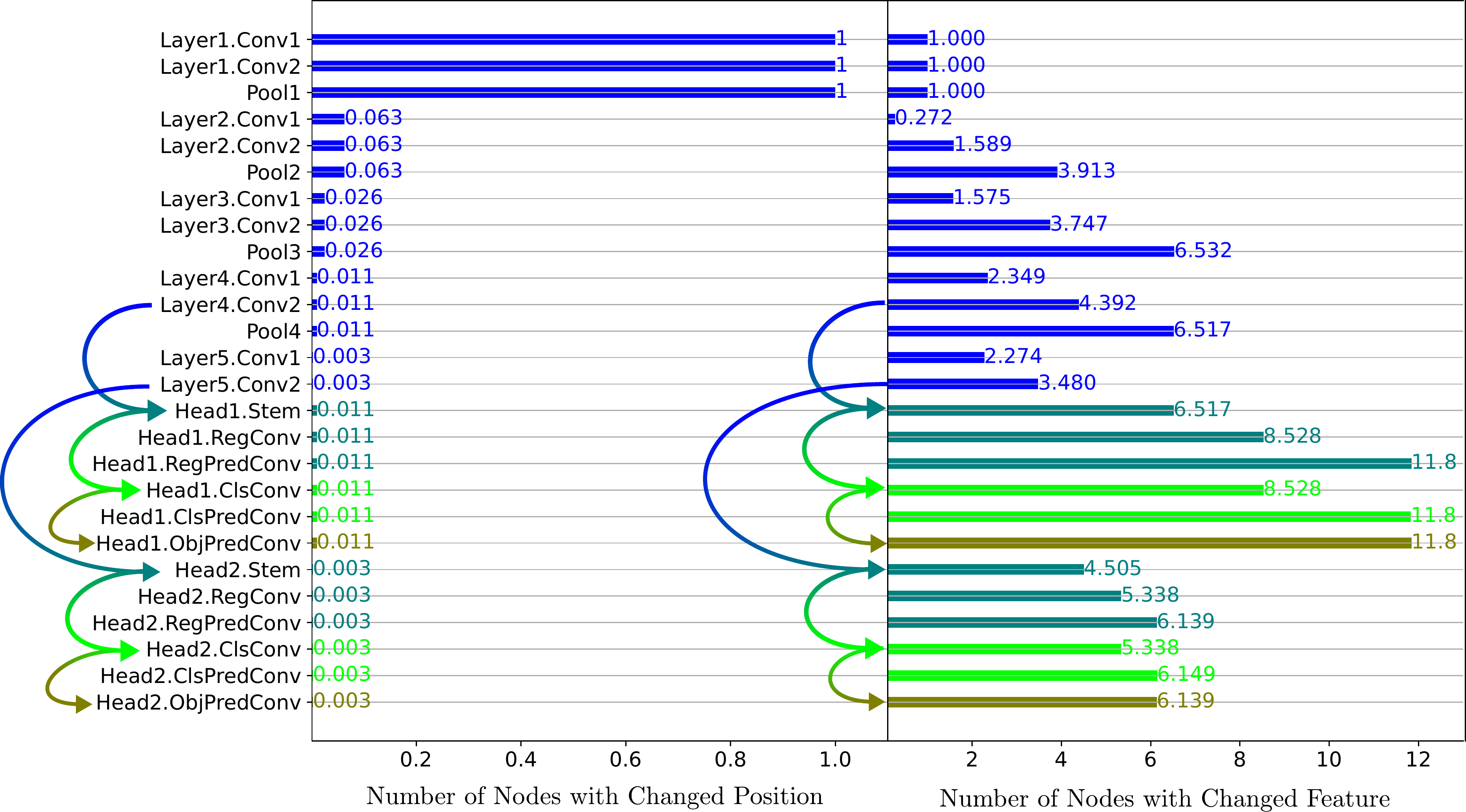}
    \end{tabular}
    \caption{Number of nodes with changed node position (left) and changed node feature (right), resulting in the green and orange update messages in Fig.~A-\ref{fig:update_propagation_detailed}. At each layer we count the number of changed nodes which enter that layer. Note the difference in x-scale. Layers on the same network branch are have the same color, and branch connections are indicated with arrows of the corresponding color.}
    \label{fig:changed_per_layer}.
\end{figure*}
\subsubsection{Per Layer FLOPS}
We analyze the number of FLOPS performed per layer, by tallying the number of FLOPS at each layer, averaged over the Gen1 dataset~\cite{Tournemire20arxiv}, as was done in Sec. M-4.1 and M-4.2 in the main manuscript. We visualize the distribution of FLOPS per layer in Fig.~A-\ref{fig:flops_per_layer}.
We can make a couple of observations: First, pooling layers contribute only a minimal amount of computation. That is because they are only concerned with updating node positions, something that only happens for few nodes, as will be seen later. Secondly, computation in the first three layers of the neural network is very small. This is because the number of nodes to be updated remains small, due to the directedness of the event graph. 
Thirdly, computation is highest in lower layers, such as the detection heads and Layer blocks 4 and 5. This is because in these layers, we need to recompute new messages for a larger number of nodes, as will be seen later. The layer with the highest computation are layers Head1.ClsConv and Head1.PredConv. These two layers have (i) the highest number of node changes (as seen later), and they also have the highest number of input and output features. Finally, output layers such as the object, regression and class predictors all have low computational complexity. This is because, despite having a high number of changed nodes, they only have few (maximally 4) output channels. Next, we will discuss the origin of these numbers by studying the number of node chagnes per layer.

\begin{figure*}
    \centering
    \begin{tabular}{c}
         \includegraphics[width=0.7\linewidth]{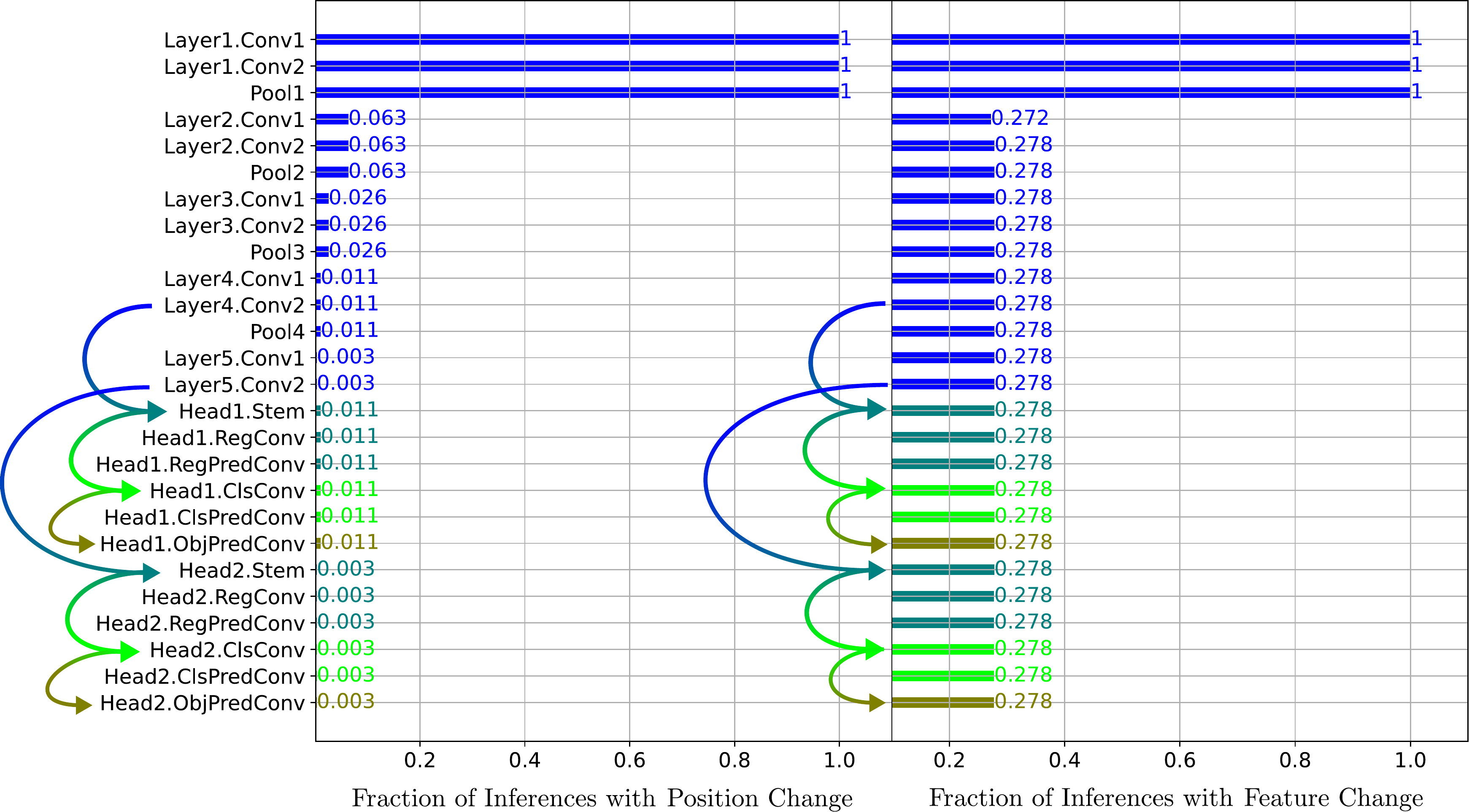}
    \end{tabular}
    \caption{Probability of there being a changed node position (left) or changed node feature (right). If both are present, an update is pruned.  At each layer we count the number of changed nodes which enter that layer. Layers on the same network branch are have the same color, and skip branch connections are indicated with arrows of the corresponding color.}
    \label{fig:filter_per_layer}.
\end{figure*}

\subsubsection{Per Layer Changed Nodes}
The above FLOPS are caused by changes in nodes features and node positions. However, it is not clear which source is dominant. To further reduce complexity, such information is vital. For this reason, we show the break-down of the number of nodes with changed features and changed positions in Fig.~A-\ref{fig:changed_per_layer} right and left respectively. At each layer we count the number of changed nodes which enter that layer. We see that most nodes changes are due to feature changes. For position changes, (left) we see that at most one node has a position change, and this corresponds to the nodes which have the new node it their voxel. We see that only 6.3\% of node insertions lead to position changes after the first pooling layer. This value diminishes to 2.6\%, 1.1\% and 0.3\% after each pooling layer. We thus conclude that most of the network computation is due to message updates of the form seen in Fig.~A-\ref{fig:update_propagation_detailed} in the bottom row. By contrast, node feature changes happen more often. After each convolution layer, the number of changes nodes increases, and after each pooling layer it decreases. We also find that most node changes happen in the high resolution detection head (Head1). This is because this head has more nodes at its disposal (a total of $g_x\times g_y$, i.e. 140 for Head1, and 35 for Head2). However, we also see that only a fraction of these nodes are updated (max 11.8). 

\subsubsection{Per Layer Probability of Filtering}
In a final step, we analyze the filtering mechanism described in Sec. M-4.2 in more detail, by investigating the probability of a node feature and node position change being present at each layer. As previously discussed. To measure this, we count for how many samples there is at least one position or feature change at each layer, and we illustrate this in Fig.~\ref{fig:filter_per_layer}. While we see the same trend for the node position as before, for the node feature it is different. Here only 27.8\% of node features are unchanged after the first pooling layer, and this number is minimally impacted in lower layers. A slight increase in the second Layer block is observed, and this has to do with the concatenation of the node position at each layer. By concatenating the node position, previously unchanged node features can experience a small update. However, as mentioned in the main text, when taking this into account, pruned updates make up $73\%$ of the total.

\subsection{Licenses}
\label{sec:app:licenses}
We use the Prophesee Gen1 dataset\cite{Tournemire20arxiv} under the ``PROPHESEE GEN1 AUTOMOTIVE DETECTION DATASET LICENSE TERMS AND CONDITIONS" found at the URL \url{https://www.prophesee.ai/2020/01/24/prophesee-gen1-automotive-detection-dataset/}.
N-Caltech101\cite{Orchard15fns} is used under the ``Creative Commons Attribution 4.0 license" and downloaded at \url{https://www.garrickorchard.com/datasets/n-caltech101}.

{\small
\bibliographystyle{ieee_fullname}
\bibliography{egbib}
}

\end{document}